# Question answering systems for health professionals at the point of care - a systematic review

Gregory Kell[1], Angus Roberts[2], Serge Umansky[3], Linglong Qian[2], Davide Ferrari[1], Frank Soboczenski[1], Byron Wallace[4], Nikhil Patel[1], Iain J Marshall[1]

[1] Department of Population Health Sciences, King's College London
[2] Department of Biostatistics and Health Informatics, King's College London
[3] Metadvice Ltd
[4] Khoury College of Computer Sciences, Northeastern University

# ABSTRACT

## objective

Question answering (QA) systems have the potential to improve the quality of clinical care by providing health professionals with the latest and most relevant evidence. However, QA systems have not been widely adopted. This systematic review aims to characterize current medical QA systems, assess their suitability for healthcare, and identify areas of improvement.

## materials and methods

We searched PubMed, IEEE Xplore, ACM Digital Library, ACL Anthology and forward and backward citations on 7th February 2023. We included peer-reviewed journal and conference papers describing the design and evaluation of biomedical QA systems. Two reviewers screened titles, abstracts, and full-text articles. We conducted a narrative synthesis and risk of bias assessment for each study. We assessed the utility of biomedical QA systems.

## results

We included 79 studies and identified themes, including question realism, answer reliability, answer utility, clinical specialism, systems, usability, and evaluation methods. Clinicians' questions used to train and evaluate QA systems were restricted to certain sources, types and complexity levels. No system communicated confidence levels in the answers or sources. Many studies suffered from high risks of bias and applicability concerns. Only 8 studies completely satisfied any criterion for clinical utility, and only 7 reported user evaluations. Most systems were built with limited input from clinicians.

## discussion

While machine learning methods have led to increased accuracy, most studies imperfectly reflected real-world healthcare information needs. Key research priorities include developing more realistic healthcare QA datasets and considering the reliability of answer sources, rather than merely focusing on accuracy.

# BACKGROUND AND SIGNIFICANCE

Despite a plethora of available evidence, health professionals find answers to only half of their questions, due to time constraints [1]. This has motivated the development of online resources to answer clinicians' questions based on the latest evidence. While scientifically rigorous information resources such as UpToDate, Cochrane, and PubMed exist, Google search remains the most popular resource used in practice [4]. General-purpose search engines like Google offer ease-of-use, but rank results according to criteria that differ from Evidence-Based Medicine (EBM) principles of rigor, comprehensiveness, and reliability [4].

To address these issues, there is burgeoning research into biomedical question answering (QA) systems [5–13]. These could rival the accessibility and speed of Google or "curbside consultations" with colleagues, while providing answers based on reliable, up-to-date evidence. Moreover, Google is free to access, while services such as service UpToDate charge for access and require manual updates. On the other hand, biomedical QA systems could be updated automatically. More recently, rapid advances in language modelling (particularly large language models [LLMs] such as GPT [14], and Galactica [15]) could allow healthcare professionals to request and receive natural language guidance summarizing evidence directly.

Many papers (e.g. [5,6,8,10,16,17]) have described the development and evaluation of biomedical QA systems. However, the majority have not seen use in practice. We explored this problem previously [18], and argue that key reasons for non-uptake include answers which are not useful in real-life clinical practice (e.g. yes/no, factoids, or answers not applicable to the locality or setting); systems that do not justify answers, communicate uncertainties, or resolve contradictions [5,6,10,16,17]. Some existing papers have surveyed the literature on biomedical question answering (e.g. [19,20]) and found that few systems explain the reasoning for the returned answers, use all available domain knowledge, generate answers that reflect conflicting sources and are able to answer non-English questions.

Our contributions are to comprehensively characterize existing systems and their limitations, with the hope of identifying key issues whose resolution would allow for QA systems to be used in practice. We focus on complete QA systems as opposed to subcomponents.

# MATERIALS AND METHODS

We conducted a systematic review and narrative synthesis of biomedical QA research, focusing on studies describing the development and evaluation of such systems. The protocol for this review is registered in PROSPERO[1] and the Open Science Framework[2].

Studies were eligible if they were: (1) published in peer-reviewed conference proceedings and journals, (2) in English language, (3) described complete QA systems (i.e. papers describing only subcomponent methods were excluded), evaluated the QA system (either based on a dataset of questions and answers, or a user study), (5) focused on biomedical QA for healthcare professionals. We excluded studies: (1) of QA systems for consumers/patients and (2) using modalities other than text, e.g., vision. We searched PubMed, IEEE Xplore, ACM Digital Library, ACL Anthology and forward and backward citations on 7th February 2023, using the following search strategy adapted for each database's syntax:

*("question answering" OR "question-answering") AND (clinic* OR medic* OR biomedic* OR health*)*

Deduplicated titles and abstracts were double screened by GK (all) and DF and LQA (50% each). Disagreements were resolved via discussion, adjudicated by IJM. The same process was followed for full texts.

We used a structured data collection form which we refined after piloting (Appendix A). We conducted a narrative synthesis following the steps recommended by Popay *et al.* [21]. Specifically, we conducted an initial synthesis by creating textual descriptions of each study and tabulating data on methods, datasets, evaluation methods, and findings, and creating conceptual maps. We assessed the robustness of findings via a risk of bias assessment, and by evaluating QA systems' suitability for real-world use.

We evaluated the suitability of QA systems for use in practice, via criteria we developed previously and introduced in our position paper [18]. This paper described how problems with transparency, trustworthiness, and provenance of health information contribute to the non-adoption of QA systems in real-world use. We proposed the following markers of high-quality QA systems. 1) Answers should come from reliable sources; 2) Systems should provide guidance where possible; 3) Answers should be relevant to the clinician's setting; 4) Sufficient rationale should accompany the answers; 5) Conflicting evidence should be resolved appropriately; and 6) Systems should consider and communicate uncertainties. We rated each system as completely, partially, or not meeting these criteria. We provide more detail regarding application of these criteria in Appendix B. Quality assessments were done in duplicate by GK (all papers), and LQ and DF (half of all papers each). Final assessments were decided through discussion and adjudicated by IJM.

---

[1] PROSPERO registration ID: CRD42021266053

[2] OSF registration DOI: 10.17605/OSF.IO/4AM8D

In the absence of a directly relevant bias tool, we adapted PROBAST for use with QA studies [22]. PROBAST evaluates study design, conduct, or analysis which can lead to biases in clinical predictive modelling studies. QA systems are like predictive models, but rather than predicting a diagnosis (based on some clinical criteria), they predict the best answer for a given question.

We adapted PROBAST to consider the quality of studies' 1) questions (analogous to *population* in the original PROBAST), 2) input features (e.g. bag-of-words, neural embeddings, etc., analogous to *predictors*), and 3) answers (analogous to *outcomes*). For each criterion, we assessed whether design problems led to *risk of bias*. We then assessed the studies for *applicability* concerns (i.e., relevance of questions, models, and answers to general clinical practice). Risks of bias and applicability concerns were rated as high, low, or unclear for each paper. We provide the modified PROBAST in the Supplementary Materials; this may be useful to other researchers assessing QA systems. Other AI-focused tools (e.g. APPRAISE-AI [23]) are rapidly becoming available; they cover similar aspects of bias to PROBAST.

We report our review according to the PRISMA [24] and SwiM guidance [25]. We provide raw data in the Supplementary Materials and present the final narrative synthesis below.

# RESULTS

The flow of studies, and reasons for inclusion/exclusion are shown in Figure 1. We included 79 of 7,506 records identified in the searches in the final synthesis. Characteristics of included studies are described in Table 1 and Figure 2.

*Figure 1: PRISMA flow diagram.*

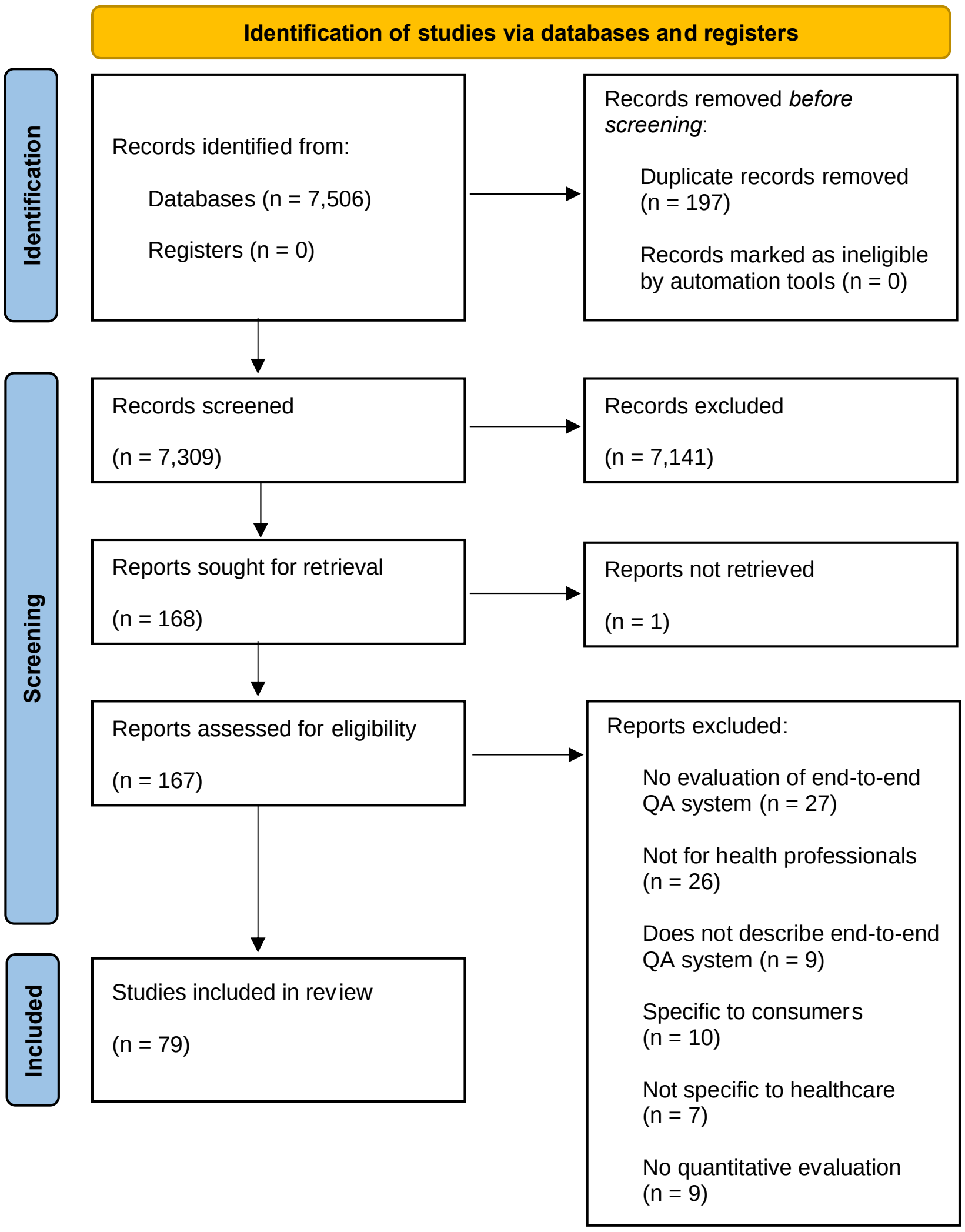

*Figure 2: Number of papers with each category of domain, method, question, answer source and answer type. The distinction was made between a major category and all the others, as one main category tended to dominate several smaller others. Table 1 contains more detail on the specifics of each paper.*

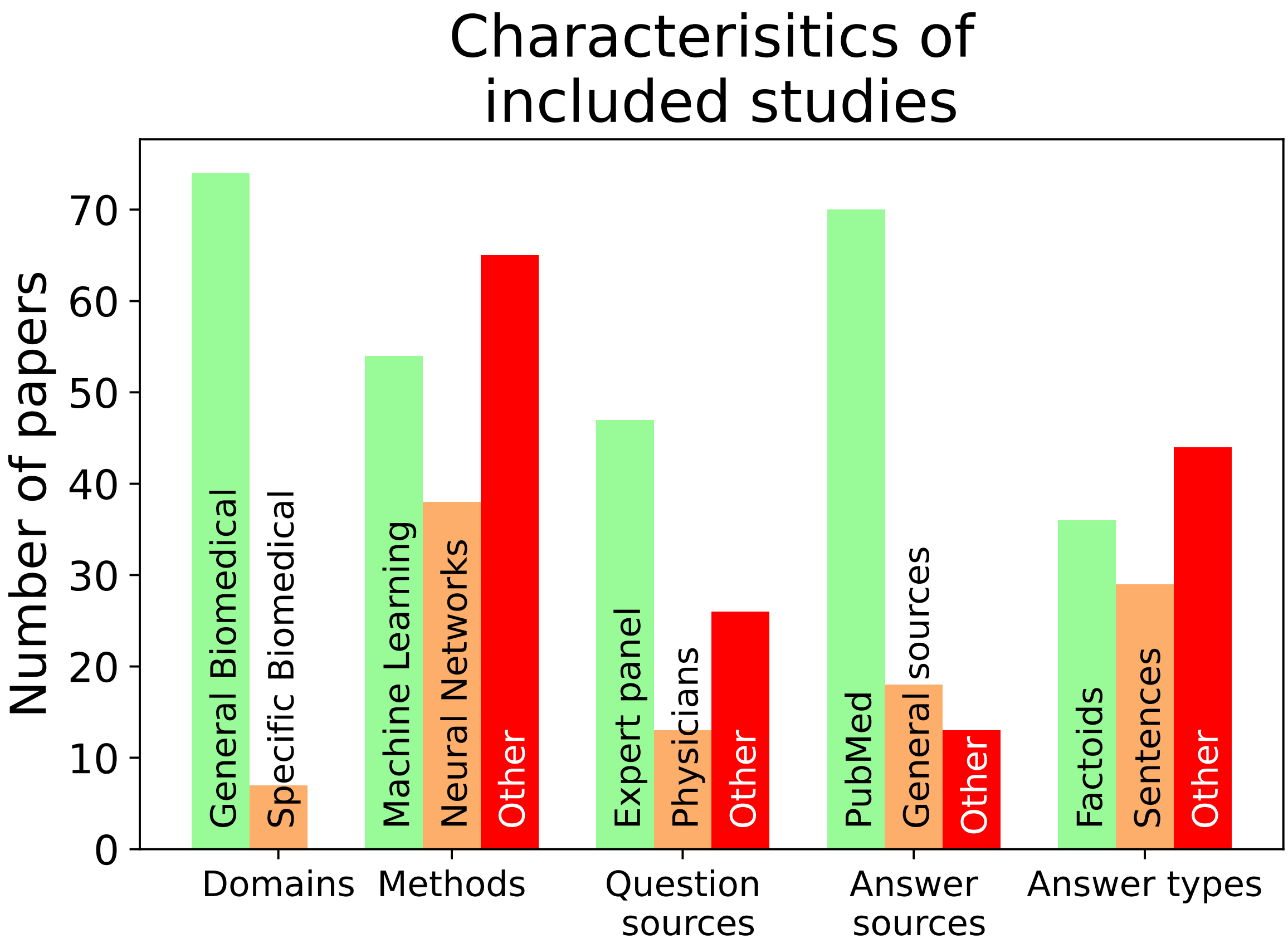

*Table 1: Characteristics of included studies.*

| **Study** | **Model/method** | **Evaluation question sources** | **Evaluation answer sources** |
|---|---|---|---|
| Demner-Fushman et al (2006) a | Semantic type classifier (UMLS, MeSH)<br>PICO classifier<br>Rule-based system<br>Machine learning system | Physicians | PubMed |
| Demner-Fushman et al (2006) b | Semantic type classifier (UMLS)<br>Clustering | Authors | PubMed |
| Lee et al (2006) | Question classification<br>Query term generation<br>TF-IDF<br>Document retrieval | Physicians | PubMed<br>World wide web |
| | Lexico-syntactic patterns | | |
| Weiming et al (2006) | Semantic type classifier (UMLS)<br>Semantic relation extraction<br>BM25<br>TF-IDF<br>Boolean search | Unclear | Medical documents |
| Demner-Fushman et al (2007) | Semantic type classifier (UMLS, MeSH)<br>PICO classifier<br>Rule-based system<br>Machine learning system | Physicians | PubMed |
| Sondhi et al (2007) | Semantic type classifier (UMLS, ICD-9)<br>Document ranking<br>Clustering | Physicians | PubMed |
| Yu et al (2007) a | User study of different systems | Physicians in practice | World wide web<br>Online dictionaries<br>PubMed |
| Yu et al (2007) b | Naïve Bayes<br>Lexico-syntactic patterns<br>TF-IDF<br>Information retrieval | Physicians in practice | World wide web<br>PubMed |
| Makar et al (2008) | Bayesian classifier<br>Part of speech tagger<br>Text extractor<br>Summarizer | Physicians in practice | Wikipedia<br>Google |
| Cao et al (2009) | BM25<br>Term frequency<br>Unique term frequency<br>Longest common subsequence | Physicians | MEDLINE<br>eMedicine documents<br>clinical guidelines<br>PubMed Central<br>Wikipedia |
| Gobeil et al (2009) | MeSH descriptors<br>Information retrieval<br>Information extraction | Authors | PubMed |
| Pasche et al (2009) a | Logical rules<br>Information retrieval | Authors | PubMed |
| Pasche et al (2009) b | Logical rules<br>Information retrieval | Authors | PubMed |
| Xu et al (2009) | Semantic type classifier (UMLS)<br>Question type classifier<br>Keyword extractor<br>Passage retrieval<br>Answer extraction | Unclear | Unclear |
| Olvera-Lobo et al (2010) | START: open-domain QA system<br>MedQA: restricted-domain QA system | Health website | START: Wikipedia<br>Merriam-Webster Dictionary<br>American Medical Association I<br>MDB<br>Yahoo<br>Webopedia.com<br><br>MedQA: MEDLINE<br>Dictionary of Cancer Terms<br>Wikipedia<br>Google<br>Dorland's Illustrated Medical Dictionary<br>Medline Plus<br>Technical and Popular Medical Terms<br>National Immunization Program Glossary |
| Tutos et al (2010) | User study on different systems | Physicians | PubMed<br>World wide web<br>Brainboost |
| Cairns et al (2011) | UMLS<br>Rule-based algorithms<br>Support vector machine | Physicians in practice | Medical wiki curated by approved physicians and doctoral-degreed biomedical students |
| Cao et al (2011) | Semantic type classifier (UMLS)<br>Related questions extraction<br>Information retrieval<br>Information extraction<br>Summarisation | Unclear | Medical documents |
| Cruchet et al (2012) | Semantic type classifier (UMLS)<br>Medical term classifier<br>Keyword-based retrieval | Physicians in practice | HONcode certified sites, e.g. WebMD, Everyday Health, Drugs.com, and Healthline |
| Doucette et al (2012) | Inference rules<br>Semantic reasoner | Synthetic patient data | Synthetic patient data |
| Ni et al (2012) | PICO classifier<br>Rules-based system<br>Template/pattern matching<br>Information retrieval<br>Machine learning system<br>Answer candidate scoring | HMedical health website | Medical health website |
| Ben Abacha and Zweigenbaum (2015) | Semantic Web<br>SPARQL<br>Semantic graphs<br>UMLS concepts<br>UMLS semantic type<br>Support vector machines<br>Conditional random fields<br>Rule-based methods | Physicians | Pubmed |
| Gobeill et al (2015) | Gene Ontology concepts<br>Lazy pattern matching<br>KNN<br>BM25<br>Information retrieval | Authors | PubMed |
| Hristovski et al (2015) | Semantic relation extraction (UMLS)<br>Semantic relation retrieval | Authors | PubMed |
| Li et al (2015) | Word2Vec<br>Markov random field | Expert panel | PubMed |
| Tsatsaronis et al (2015) | Comparison of different systems on the BioASQ dataset | Expert panel | PubMed |
| Vong et al (2015) | PICO classifier<br>Clustering | Authors | PubMed |
| Goodwin et al (2016) | Knowledge graph<br>Conditional random fields<br>Bayesian inference | Unclear | Electronic health records<br>PubMed |
| Yang et al (2016) | Logistic Regression, Classification via Regression, Simple Logistic | Expert panel | PubMed |
| Brokos et al (2016) | TF-IDF<br>Word mover's distance | Expert panel | PubMed |
| Krithara et al (2016) | Comparison of different systems on the BioASQ dataset | Expert panel | PubMed |
| Sarrouti and El Alaoui (2017) | UMLS concepts<br>BM25 | Expert panel | PubMed |
| Sarrouti et al (2017) | UMLS<br>BM25 | Expert panel | PubMed |
| Jin et al (2017) | Bag of words<br>Term frequency<br>Collection frequency<br>Sequential dependence models<br>Divergence from randomness models<br>Multimodal strategies | Expert panel | PubMed |
| Neves et al (2017) | Question processing (regular expressions, semantic types, named entities, keywords),<br>Document/passage retrieval,<br>Answer extraction | Expert panel | PubMed |
| Wiese et al (2017) a | RNN<br>Domain adaptation | Expert panel | PubMed |
| Wiese et al (2017) b | RNN<br>Domain adaptation | Expert panel | PubMed |
| Nentidis et al (2017) | Comparison of different systems on the BioASQ dataset | Expert panel | PubMed |
| Du et al (2018) | GloVe<br>LSTM<br>Self-attention | Expert panel | PubMed |
| Eckert et al (2018) | Semantic role labelling | Expert panel | PubMed |
| Papagiannopoulou et al (2018) | Binary relevance models<br>Linear SVMs,<br>Labelled LDA variant<br>Prior LDA<br>Fast XML<br>HOMER-BR<br>Multi-label ensemble | Expert panel | PubMed |
| Dimitriadis et al (2019) | Word2Vec<br>WordNet<br>Custom textual features<br>Logistic regression<br>Support vector machine<br>XGBoost | Expert panel | PubMed |
| Du et al (2019) | GloVe<br>LSTM<br>Self-attention<br>Cross-attention | Expert panel | PubMed |
| Jin et al (2019) | BioBERT | Titles of papers | PubMed |
| Oita et al (2019) | Dynamic Memory Networks<br>Bidirectional Attention Flow<br>Transfer learning,<br>Biomedical named entity recognition<br>Corroboration of semantic evidence | Expert panel | PubMed |
| Ozyurt et al (2019)] | GloVe<br>BERT<br>Inverse document frequency<br>Relaxed word mover's distance | Expert panel | PubMed |
| Jin et al (2019) | TF-IDF<br>Noun extraction<br>Part of speech tagger<br>Semantic type classifier (UMLS) | Expert panel | PubMed |
| | Query expansion (MeSH)<br>Markov random field<br>Divergence from randomness<br>Model ensemble | | |
| Wasim et al (2019) | Rules-based system<br>Semantic type classifier (UMLS)<br>Logistic regression | Expert panel | PubMed |
| Du et al (2020) | BERT<br>BiLSTM<br>Self-attention | Expert panel | PubMed |
| Yan et al (2020) | Binary classification<br>RNNs<br>Semi-supervised learning<br>Recursive autoencoders | Expert panel | PubMed |
| Kaddari et al (2020) | Survey of existing models | Expert panel | PubMed |
| Nishida et al (2020) | BERT<br>Domain adaptation<br>Multi-task learning | Expert panel<br>Crowdworkers | PubMed<br>Wikipedia |
| Omar et al (2020) | Convolutional neural networks<br>Attention<br>Gated convolutions<br>Gated attention | PubMed | PubMed |
| Ozyurt et al (2020) a | GloVe<br>BERT<br>Inverse document frequency<br>Relaxed word mover's distance | Expert panel | PubMed |
| Ozyurt et al (2020) b | ELECTRA | Expert panel | PubMed |
| Sarrouti et al (2020) | Lexico-syntactic patterns<br>Support vector machine<br>Semantic type classifier (UMLS)<br>TF-IDF<br>Semantic similarity-based retrieval<br>BM25 | Expert panel | PuMed |
| | Sentiment analysis | | |
| Shin et al (2020) | BioMegatron | Expert panel | PuMed |
| Wang et al (2020) | Event extraction<br>SciBERT | Authors | PubMed |
| Alzubi et al (2021) | TF-IDF<br>BERT | Authors | PubMed |
| Du et al (2021) | QANet<br>BERT<br>GloVe<br>Model weighting | Expert panel | PubMed |
| Nishida et al (2021) | BERT<br>fastText | Expert panel<br>Crowdworkers | PubMed<br>Wikipedia |
| Peng et al (2021) | BERT<br>BiLSTM<br>Bagging | Expert panel | PubMed |
| Pergola et al (2021) | BERT<br>Masking strategies | Epidemiologists<br>Medical doctors,<br>Medical students,<br>Expert panel | PubMed<br>World Health Organization's Covid-19 Database<br>Preprint servers |
| Wu et al (2021) | BERT<br>Numerical encodings | Expert panel<br>PubMed | PubMed |
| Xu et al (2021) | BERT<br>Syntactic and lexical features<br>Feature fusion<br>Transformer | Expert panel | PubMed |
| Bai et al (2022) | Dual-encoder<br>BioBERT | Expert panel | PubMed |
| Du et al (2022) | QANet<br>BERT<br>GloVe<br>Model weighting | Expert panel | PubMed |
| Kia et al (2022) | Convolution neural network<br>Attention | Authors | PubMed |
| Naseem et al (2022) | ALBERT | Expert panel | PubMed |
| Pappas et al (2022) | ALBERT-XL | Expert panel | PubMed |
| Raza et al (2022) | BM25<br>MPNet | Expert panel | PubMed |
| Rakotoson et al (2022) | BERT<br>RoBERTa<br>T5<br>Boolean classifier | Expert panel<br>PubMed | PubMed |
| Wang et al (2022) | Event extraction<br>SciBERT<br>Domain adaptation | Authors | PubMed |
| Weinzierl et al (2022) | BERT<br>BM25<br>Question generation<br>Question entailment recognition | Expert panel | PubMed |
| Yoon et al (2022) | BERT<br>Sequence tagging<br>BiLSTM-CRF | Expert panel | PubMed |
| Zhang et al (2022) | BERT<br>BM25 | Expert panel | PubMed |
| Zhu et al (2022) | BERT<br>RoBERTa<br>T5<br>XGBoost | PubMed | PubMed |
| Bai et al (2023) | Knowledge distillation<br>Adversarial lLearning<br>BioBERT | Expert panel | PubMed |
| Raza et al (2023) | BM25<br>MPNet | Expert panel | PubMed |

## risk of bias, applicability, and utility

We summarise the risks of bias in Figure 3; individual study assessments are in the Supplementary Materials. 85% of systems had a high risk of bias overall; primarily driven by problems in the questions used to develop and evaluate the systems. Many studies used unrealistically simple questions or covered too few information needs for a general biomedical QA system. Most questions were hypothetical, and not generated by health professionals.

*Figure 3: Number of papers achieving each risk of bias and applicability concern classification. Risk of bias refers to the risk of a divergence between the stated problem the paper tries to solve and the execution for reasons such as an unrealistic dataset or failing to split data for training and evaluation. Applicability refers to how applicable the system is to the review.*

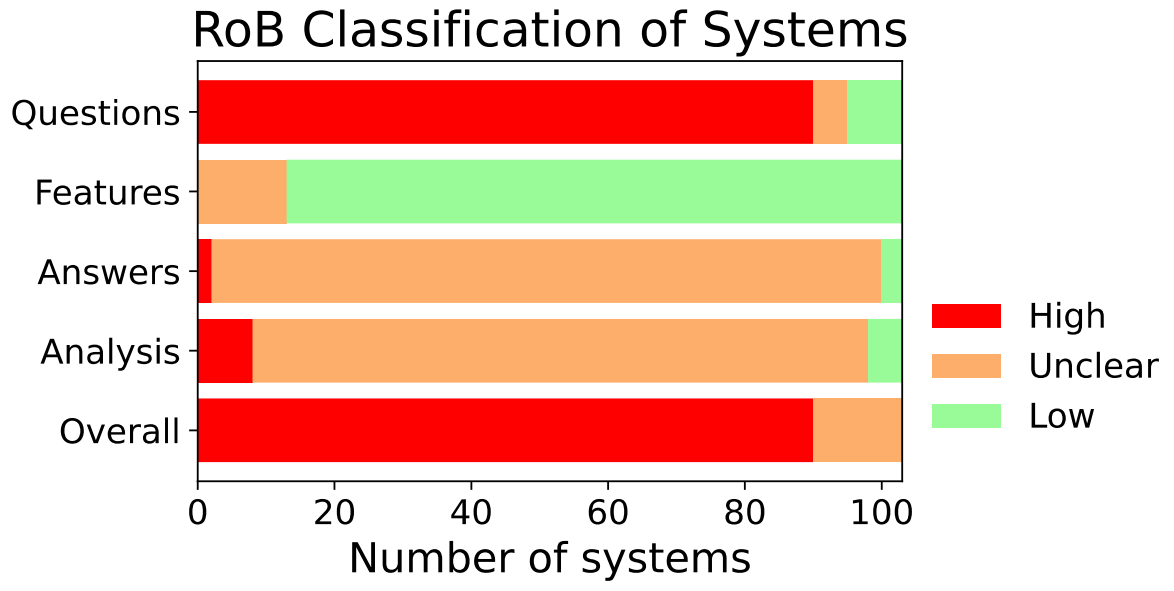

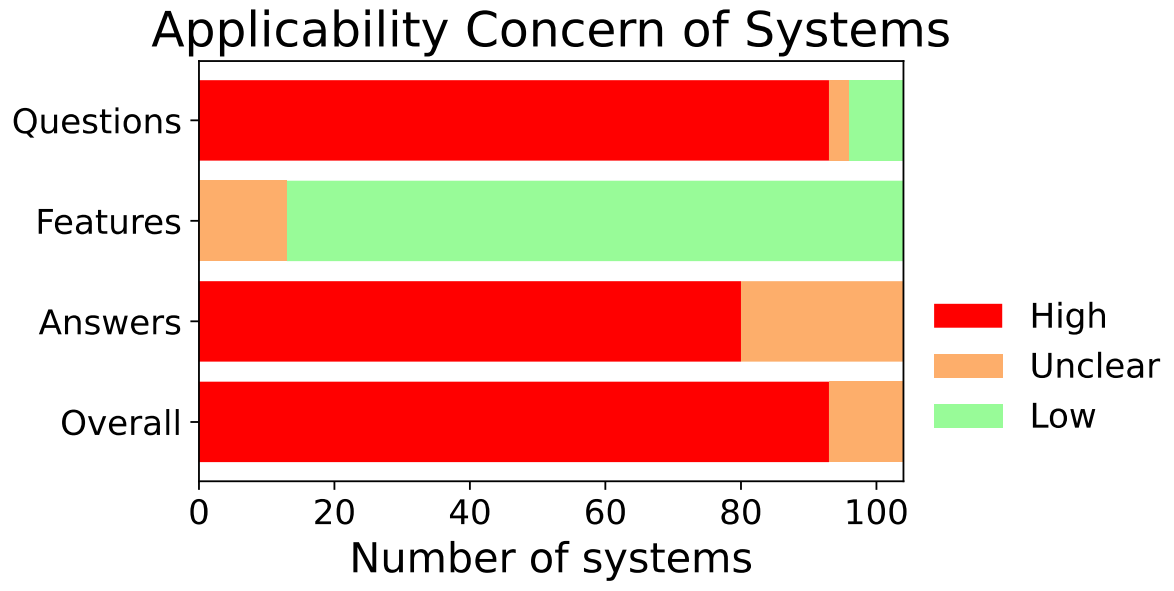

Most systems were at low risk of bias for defining and extracting machine learning (ML) features (e.g., deciding on predictive features without reference to the reference answers). Most studies did not provide clear descriptions of answer data or evaluation methodology (e.g., details about the source of answers) which led to unclear risk of bias assessments for most papers' answers. Additionally, no answer was relevant to the biomedical QA domain. This led to high applicability concerns for most papers.

We present utility scores in Figure 4. Few systems completely met any criterion. Two systems [26,27] provided rationales (i.e., justifications and sources) for their answers; five systems were judged to use reliable sources [11,28–31]; one system resolved conflicting information [26] and one system communicated uncertainties [26]. Very few systems provided contextually relevant answers (i.e., locality-specific information, or specialty), while most systems provided clinical guidance at least partially (rather than basic science or less actionable information).

*Figure 4: Number of papers achieving each satisfaction classification for each criterion.*

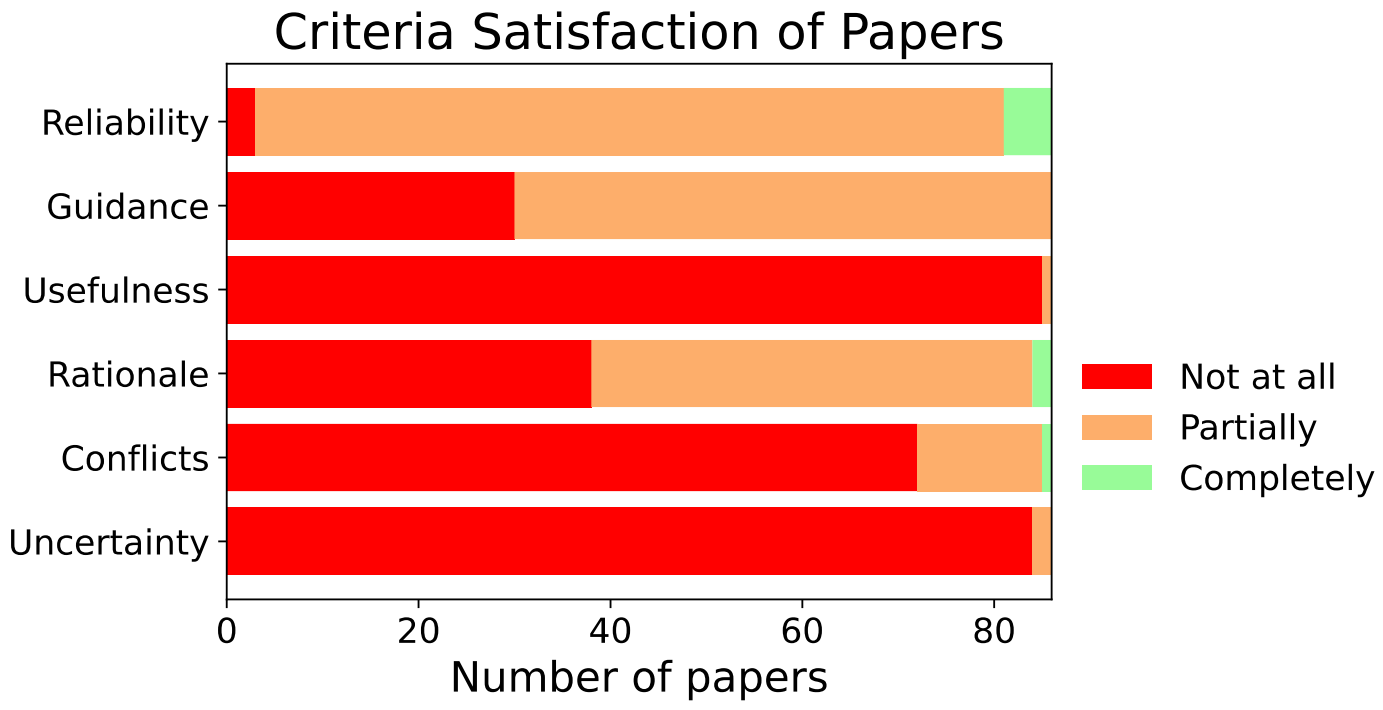

## computational methods

Most QA systems used a knowledge base (i.e. database of answers) that was created using documents from PubMed or other medical information sources (see Figure 5 for a typical example, from Alzubi et al. [32])). Documents were either stored in structured form (knowledge graphs or RDF triples) or as unstructured texts.

*Figure 5: Typical QA architecture as used by Alzubi et al. [32]*

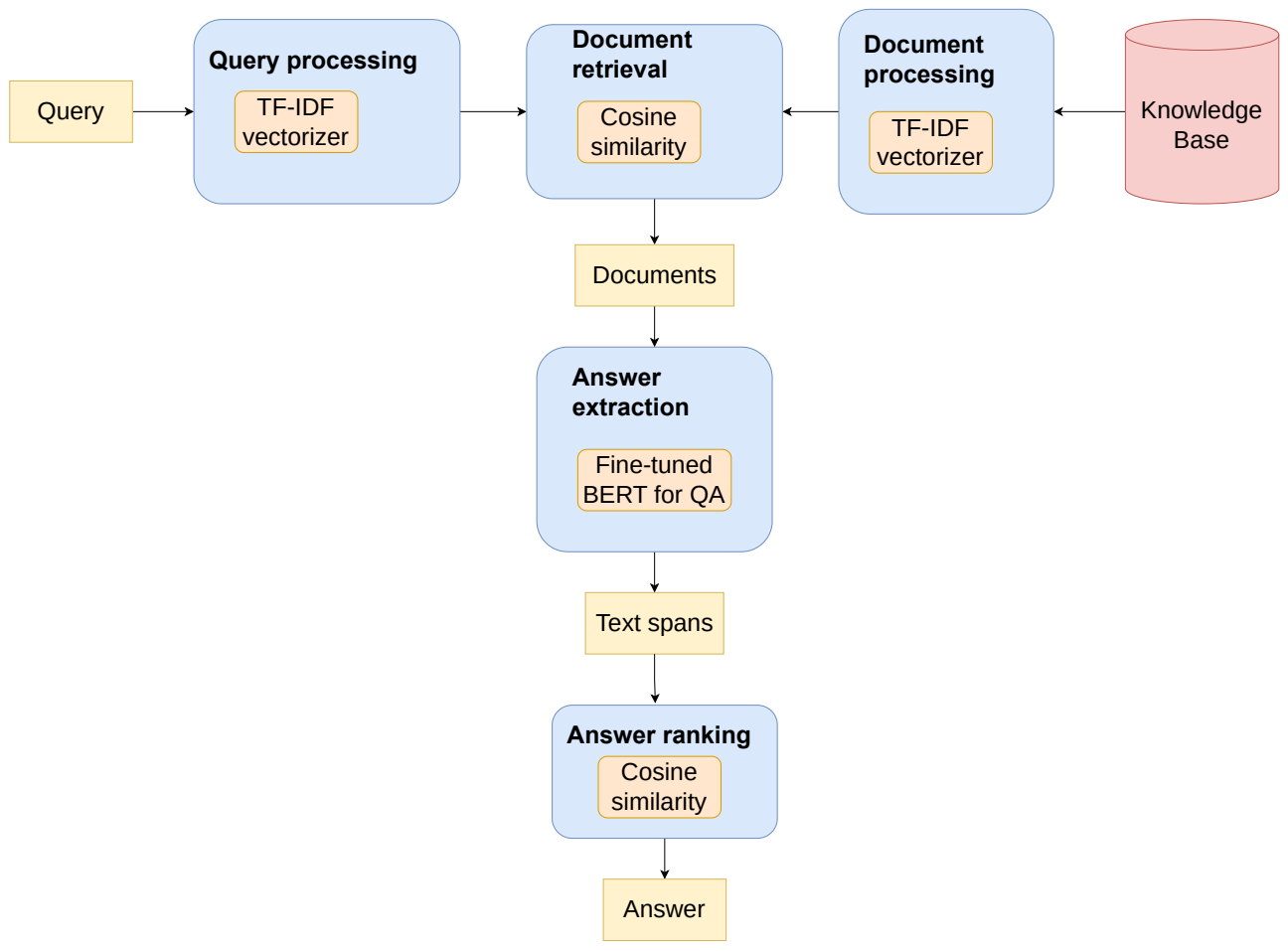

For a given user query, the system would retrieve the most relevant answer(s) from the knowledge base. Knowledge graph-based (KG) (1 study), neural (24 studies) and modular systems (39 studies) were evaluated in the included studies (see Figure 2 and Appendix C). KG-based systems accept natural language questions and convert them to KG-specific queries (e.g. Cypher queries [33]). Modular systems comprise several distinct components (e.g., question analysis, document retrieval, answer generation) designed separately and combined to form a QA system. Neural systems can be modular or monolithic.

All studies made use of datasets of questions with known answers. These datasets were used to train ML models (e.g. document retrieval and answer extraction) and evaluate system performance. The topic focus of these datasets dictates the area(s) for which the QA can be successfully used; the quality of these datasets impacts both the accuracy of trained models and the reliability of the evaluations.

With regards to neural systems, 9 studies ([32,34–41]) incorporated pretrained LLMs (e.g. BERT [42], BioBERT [43] and GPT [44]) in their QA pipelines for text span extraction, sentence reranking and integrating sentiment information. These models were used to find potential answer text spans given questions and passages. Four studies ([27,36,37,40]) found fine-tuning pretrained LLMs on biomedical data led to improvements in performance compared with only using only a general-domain LLM. No experiments were conducted on LLMs that were trained only on biomedical data.

Few studies used common datasets for training or evaluation. However, several of the included studies arose from the BioASQ 5b [45] and 6b [46] shared tasks, which aimed to answer four types of questions (yes/no, factoids, list, and summary questions) and had two phases: information retrieval and exact answer production. Three studies arising from BioASQ ([54,55,66]) evaluated QA systems with a neural component, while five studies ([53–55,58,66]) evaluated QA systems that relied only on rule-based or classical ML components (e.g. support vector machines). The neural components encoded questions and passages with a recurrent neural network (RNN) that were then used to create intermediate representations before answers were generated with additional layers. Comparing results across the BioASQ studies suggests generally that QA systems employing ML components outperformed those that relied solely on rule-based components (see Figure 6 and Appendix C).

*Figure 6: Results of the BioASQ 5b and 6b challenges for factoid-type answers.*

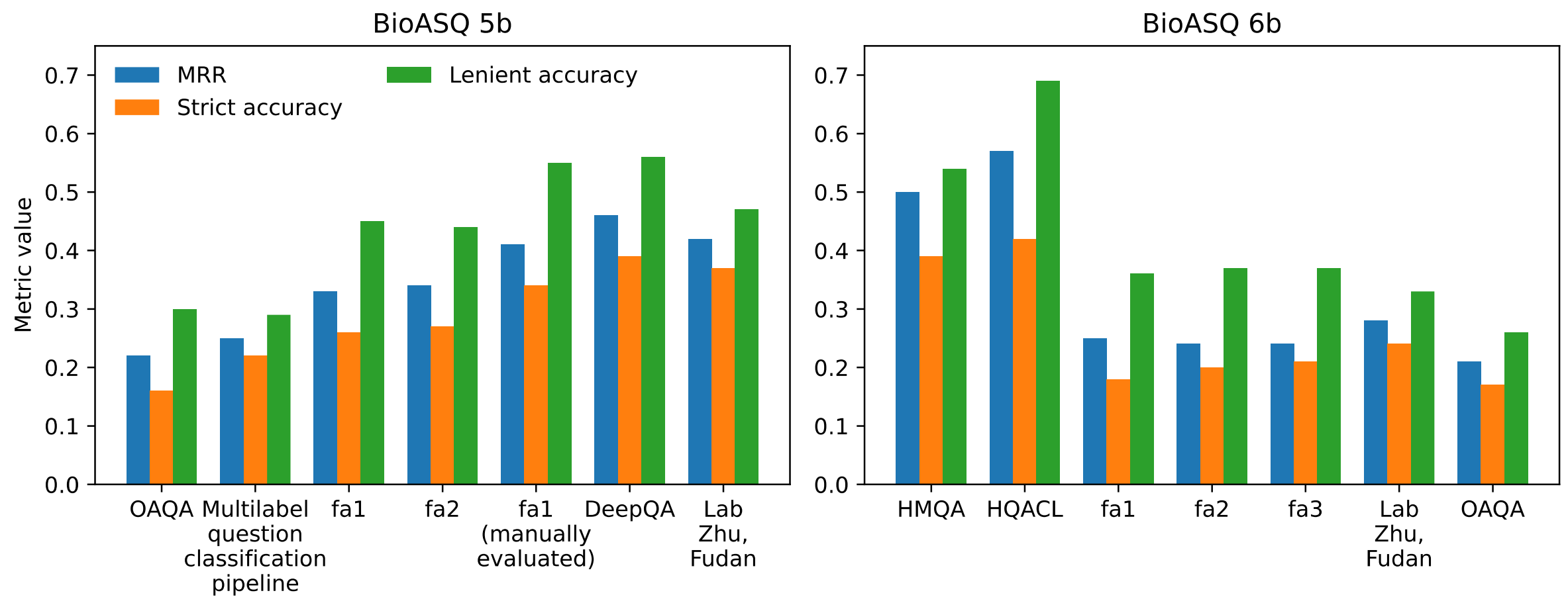

Two papers included a numerical component in their QA pipelines. For example, one paper ([27]) used numerical results (e.g., odds ratios from clinical trial reports) to generate answers either to answer statistical questions (i.e. “Do preoperative statins reduce atrial fibrillation after coronary artery bypass grafting?”). One study ([27]) generated BERT-style embeddings using both textual and numerical encodings, leading to improved performance compared with using text alone.

## topic areas

53 studies ([5,8,11,16,17,26–31,34,36–41,47–81]) described QA systems covering a wide breadth of biomedical topics (Figure 2). These systems typically sourced answers from the unfiltered medical literature (e.g., PubMed, covering both clinical practice guidelines and primary studies, including laboratory science and epidemiology). Eight studies examined specific specialties: one study focused on bacteriotherapy [82], two focused on genetics/genomics [72,83], and 5 on Covid-19 [32,78,84–86]. The genomics and Covid-19 systems were designed for specialists, while the bacteriotherapy system generated rules for managing antibiotic prescribing via a QA interface.

## question datasets

Studies used several sources to generate question datasets (see Figure 2 and Appendix D). We group these into questions collected from health professionals (either collected in the course of work or elicited as generate hypothetical questions; 14 studies), those generated by topic experts (13 studies), people without direct healthcare experience (e.g., crowdworkers; three studies), and automatically/algorithmically derived (scraped from health websites, or generated from abstract titles; two studies). In nine papers, questions were written by study authors.

Only 5 [17,48,51,61,81] studies used genuine questions posed by clinicians during consultations. Two studies ([11,28]) used either simple or simplified questions. Examples of simple questions include “How to beat recurrent UTIs?” [28] and “What is the best treatment for analgesic rebound headaches?” [11]. Questions the BioASQ challenge questions [53] were created by an expert panel. BioASQ questions were restricted to yes/no, factoid and summary-type questions, and tended to have a highly technical focus. For example, the question “Which is the most common disease attributed to malfunction or absence of primary cilia?” could be answered with a factoid: "autosomal recessive polycystic kidney disease”. Alternatively, it could be answered with a summary (see Appendix E for example). One study included definition questions created by the authors ([71]), while another ([32]) included author-created factoid-style questions about a particular topic. Two studies ([28] [29]) utilized questions derived from health websites: one included questions generated by physicians([28]) and one ([29]) used questions that were of unclear provenance.

While biomedical question sources enabled training of models, general domain QA datasets created using crowdworkers (e.g. SQuAD [87,88]) were used to pretrain QA models in 3 studies ([36,63,64]). These pretrained models were then fine-tuned on biomedical QA datasets (e.g. BioASQ [53]). Pretraining QA models on general crowdworker-created datasets prior to fine-tuning on biomedical datasets led to overall improvements in model performance in all 3 studies that explored this approach. In other words, pretraining on general-domain data led to an improvement in performance compared with training only on biomedical data.

## reliability of answer sources

The answer sources used by the studies are summarized in Figure 2 and Appendix G. Two studies ([11,30]) found ranking biomedical articles by strength of evidence (based on publication types, source journals and study design) improved accuracy (e.g. precision at 10 documents, mean average precision, mean reciprocal rank). None of the other studies accounted for differences in answer reliability within datasets (i.e. information from major guidelines was treated equally to a letter to the editor).

Several studies included answers derived from health websites such as Trip Answers [29], WebMD [71], HON-certified websites [73], clinical guidelines and eMedicine[3] documents. These answers were created by qualified physicians and underwent a review process. On the other hand, 3 studies ([48], [57], [71]) explored systems that provided only term definitions from medical dictionaries. One study derived answers entirely from general domain sources ([28]), while another generated answers from a combination of medical and general sources. In the case of the latter, only the medical sources had a rigorous validation process ([71]). Two QA systems[29,73] only derived the answers from health websites containing information that was vetted by the administrators. One study found that restricting the QA document collection based on trustworthiness increased the relevance of answers ([73]).

[3] https://emedicine.medscape.com/

## detail of answers

Systems we reviewed varied in terms of what they produced as an 'answer' (Figure 2 and Appendix H). Answers consisting of only of one word (i.e., cloze-style QA), factoids (a word or phrase, e.g., aspirin 3g), list of factoids, or definitions were absolute in nature and therefore did not contain guidance (Appendix H). On the other hand, contextual texts (e.g., ideal answers [53] and document summaries [16]) that accompanied absolute answers (e.g., factoids _) may have contained guidance. Similarly, biomedical articles accompanying answers consisting of medical concepts may have also included guidance, along with the sentences accompanying yes/no/unclear answers (see Appendix H).

Several systems used a *clustered* approach to display answers. These systems grouped several candidate answers either by keyword or topics, e.g., articles/sentences about heart conditions as one cluster. Clustered answers returned by the systems in 6 studies ([5,30,54,55,61,70]) may contain guidance as the clusters are based around sentences, extracts of documents, or conclusions of abstracts. Other types of answers included abstracts and single/multiple sentences, documents and webpages and URL-based answers (Appendix H).

## evaluation

Most studies (47) considered the accuracy of answers provided (see Table 2). Some assessed the degree to which the words in the answer match the reference, i.e. accuracy, precision, recall, F1 with respect to words (e.g. ROUGE) or correct entire answers (e.g. yes/no or factoids), numbers of answers/questions, exact matches. While ROUGE [89] or BLEU [90] may quantify the degree of similarity between candidate answers and the reference sentence, they are unable to account for e.g. negation or re-phrasings. Other systems were retrieval-based and so evaluated using the position of the correct answers in the returned list (i.e., reciprocal rank, MAP, normalized discount cumulative gain). Of the models that assessed accuracy/correctness, 31 used internal cross-validation, while 17 were evaluated on an independent dataset. Only 7 studies evaluated their design, system usability, or the relevance of the answer to the question as assessed by users. The most popular answer source was PubMed; most systems used a single source of answers.

*Table 2: Grouping of papers according to accuracy metric.*

| Metric | Metric type | Papers | Number of papers |
|---|---|---|---|
| Accuracy | Accuracy/correctness | [16,36,37,39–41,47,49,51,58,59,64,66,70,74–76,91,93,95–97,99,104–107,110,114] | 29 |
| Precision | Accuracy/correctness | [6,11,12,16,26,29,34,41,53,55,56,58,59,66,80,81,94–99,104,105,108,111–113] | 28 |
| Recall | Accuracy/correctness | [16,26,41,53,55,59,66,71,80–83,94–97,99,104,105,108,111–113] | 24 |
| Reciprocal rank | Accuracy/correctness | [6,8,12,12,16,34,36–40,59,63,64,66,71,72,74,75,80,95–97,99,101–108,110] | 32 |
| F1 | Accuracy/correctness | [16,26,29,41,53,59,63,64,66,77,79,81,84–86,91,94,96,97,99,101–106,108–110,112–114] | 32 |
| ROUGE | Accuracy/correctness | [16,26,31,53,91,96,97,99,101,104] | 10 |
| Time taken to find answer | Usability | [5,17,26,28,48,51] | 6 |
| Likert score | Usability | [5,17,28,30,48,57,61] | 7 |
| Action frequency | Usability | [17] | 1 |
| MAP | Accuracy/correctness | [53,56,62,72,92,94,100,101,105] | 9 |
| Numbers of queries/answers | Accuracy/correctness | [70–72,101,112] | 5 |
| Exact matches | Accuracy/correctness | [26,32,77,79,84–86,108–110] | 10 |
| Normalized discount cumulative gain | Accuracy/correctness | [78,98] | 2 |
| AUC ROC | Accuracy/correctness | [12] | 1 |

## presentation and usability

Only 13 studies evaluated 7 systems that provided a user interface for user queries. These systems were MedQA [17,28,48,71,115], Omed [49], the system introduced in [51], EAGLi [56,82,83], AskHERMES [5,30,61], CQA-1.0 [30] and CliniCluster [30]. User interfaces are essential for assessing the performance of the systems with genuine users.

The only usability study ([30]) assessed the effectiveness of a system that clustered answers to drug questions by I (intervention) and C (comparator) elements. The answers were tagged with P-O (patient-outcome) and I/C (intervention/comparator) elements (see Appendix I for details). The participants agreed that the clustering of the answers helped them find answers more effectively, while more of the older participants found the P-O and I/C useful for finding relevant documents. Additionally, possessing prior knowledge about a given subject assisted with additional learning.

The ease of use of QA and IR systems was assessed in 3 studies ([5] [49] [17]). The systems evaluated included Google [5,17,49], MedQA [17,49], Onelook [17,49], PubMed [17,49], UpToDate [5] and AskHermes [5]. Both Doucette et al. [49] and Yu et al. [17] rated Google as being the easiest to use, followed by MedQA, Onelook and PubMed. On the other hand, Cao et al. [5] rated Google, UpToDate, and AskHermes equally in terms of ease of use.

None of the included systems presented any information about the certainty of answers; although nearly all systems used quantitative answer scoring to select the chosen answer.One study [61] evaluated two approaches to presenting answers on the AskHermes system [5]: passage-based (collection of several sentences) and sentence-based. The study found that passage-based approaches produced more relevant answers as rated by clinicians.

# DISCUSSION

We systematically reviewed studies of the development and evaluation of biomedical QA systems, focussing on their merits and drawbacks, evaluation and analysis, and the overall state of biomedical QA. Most of the included studies had high overall risks of bias and applicability concerns. Few of the papers satisfied any of utility criteria [18].

Several studies highlight obstacles that should be overcome and measures that should be taken before deploying biomedical QA systems. For example, one general-domain QA user study [93] found that users tended to prefer conventional search engines as they “felt less cognitive load” and “were more effective with it” than when they queried QA systems.

We note that commercial search engines are likely to benefit from comparatively vast development resources, and a focus on user experience. By contrast, the academic research we found tended to focus on the underlying computational methodology/models, with little attention to the user interface or experience—aspects which are likely highly influential in how QA systems are used.

Law et al. [94] found that presenting users with causal claims and scatterplots could lead users to accept unfounded claims. Nonetheless, warning users that “correlation is not causation” led to more cautious treatment of reasonable claims. Additionally, Schuff et al. [96] and Yang et al. [95] explored metrics for assessing the quality of the explanations:

answer location score (LOCA) and the Fact-Removal score (FARM), F1 score and exact matches.

More recently, there has been rapid development in LLMs, such as GPT [14], PaLM [21] and Med-PaLM [23], which are the current state-of-the-art in natural language processing. There were 9 studies included that used LLMs, but they were used for text span extraction, sentence reranking and integrating sentiment information. A nascent application of LLMs is direct summarization of one or more sources. While LLMs can produce fluent answers to any given question [98], they are vulnerable to "hallucinating" plausible but fabricated information [99–101]. This may be especially risky in healthcare due to the potentially life-threatening ramifications. One solution might be retrieval-augmented methods (where LLMS only use documents of known provenance). LLMs should be rigorously assessed before deployment in biomedical QA pipelines. This would ensure that the references provided by LLMs are genuine and that information is faithfully reproduced.

Barriers to adoption have been studied in detail in related technologies (e.g. Clinical Decision Support Systems [CDSS]). Greenhalgh et al.[124,125] introduced the NASSS framework to characterise the complex reasons why technologies succeed (or fail) in practice; finding that aspects such as the dependability of the underlying technology and organisations' readiness to adopt the new systems are critical. Similarly, Cimino and colleagues [126] found that design issues (e.g. time taken to answer each question, or the number of times a given link is clicked) were critical. We would argue that future QA research should take a broader view of evaluation if QA is to move from an academic computer science challenge to real-world benefit.

To our knowledge, this is the first systematic review of QA systems in healthcare. While other (non-systematic) reviews provide an overview of the biomedical QA field [19,20], we have evaluated existing systems and datasets for their utility in clinical practice. Furthermore, the inclusion of quantitative evaluations allowed for comparisons between different system types. Examination of questions, information sources and answer types has allowed identification of factors that affect adherence to the criteria defined in [18].

Most of the included studies were method papers describing systems that were built by computer scientists with limited input from clinicians. These systems were designed to perform well on benchmark datasets, such as BioASQ. While the studies were rigorous in their evaluation, they did consider how the systems could be used in practice. Future work should focus on translating biomedical QA research into practice.

One weakness is that we did not include purely qualitative evaluations. This might be a worthwhile SR to do in the future. We limited our search to published systems; therefore, this review would not have included any deployed systems which were not published; or systems described only in the 'grey' literature (e.g. pre-prints, PhD theses, etc). We also did not search all the CDSS literature for pipelines incorporating QA systems. Deployment of such systems might not be described in the literature, as health providers may not have provided the results. Although we would expect most relevant papers to be published in English, there may have been pertinent non-English language papers that were missed.

## implications for research

Studies to date have too often used datasets of factoids/multiple choice questions, which do not resemble real-life queries. There is a need for high quality datasets derived from real clinical queries, and actionable high quality clinical guidance.

Future research should move beyond maximising accuracy of a model alone, and include aspects of transparency, answer certainty, and information provenance (is the reliability and source of answers understood by users?). These aspects will only become more important with the advent of LLMs, which tend to generate highly plausible and fluent answers, but are not always correct.

## implications for practice

The performance of QA systems on biomedical tasks has increased over time, but the tasks are unrealistically simple. We recommend that practitioners exercise caution with any QA system which advertises accuracy only. Instead, systems should produce verifiable answers of known provenance, which make use of high-quality clinical guidelines and research.

# CONCLUSIONS

In this review we reviewed the literature on QA systems for health professionals.  Most studies assessed the accuracy of the systems on various datasets; only thirteen evaluated the usability of the systems. Few studies examined the use of the in practice and instead compared systems using biomedical QA benchmarks such as BioASQ. Although none of the included studies described systems that completely satisfied our utility criteria, they discussed several characteristics that could be appropriate for future systems. These included, limiting the document collection to reliable sources, providing more verbose answers, clustering answers according to themes/categories and employing methodologies for numerical reasoning. While an increase in the performance of QA systems on biomedical tasks has been observed over time, the tasks themselves are unrealistic. Thus, more realistic and complex datasets should be developed.

# FUNDING

This work is supported by: National Institutes of Health (NIH) under the National Library of Medicine, grant R01-LM012086-01A1, "Semi-Automating Data Extraction for Systematic Reviews". GK is a supported by a studentship funded by King's College London and Metadvice; there is no grant number associated with this funding.

# DATA AVAILABILITY

The data underlying this article are available in the article and in its online supplementary material.

# CONTRIBUTORSHIP STATEMENT

**Gregory Kell**: Conceptualisation, Data Curation, Formal Analysis, Investigation, Methodology, Visualisation, Writing – original draft. **Linglong Qian**: Formal Analysis, Investigation, Writing – review and editing. **Davide Ferrari:** Formal Analysis, Investigation, Writing – review and editing. **Frank Soboczenski:** Formal Analysis, Investigation, Writing – review and editing. **Byron Wallace:** Writing – review and editing. **Angus Roberts:** Writing – review and editing. **Serge Umansky:** Writing – review and editing. **Nikhil Patel:** Formal Analysis, Investigation, Writing – review and editing. **Iain J Marshall:** Conceptualisation, Data Curation, Formal Analysis, Investigation, Methodology, Visualisation, Writing – review and editing, Supervision.

# CONFLICT OF INTEREST STATEMENT

None declared

# APPENDIX

Contents:

# REFERENCES

1. Del Fiol G, Workman TE, Gorman PN. Clinical Questions Raised by Clinicians at the Point of Care: A Systematic Review. JAMA Internal Medicine. 2014 May;174(5):710–8.

2. Bastian H, Glasziou P, Chalmers I. Seventy-Five Trials and Eleven Systematic Reviews a Day: How Will We Ever Keep Up? PLOS Medicine. 2010 Sep 21;7(9):e1000326.

3. Hoogendam A, Stalenhoef AFH, Robbé PF de V, Overbeke AJPM. Answers to questions posed during daily patient care are more likely to be answered by UpToDate than PubMed. Journal of medical Internet research. 2008 Oct;10(4):e29–e29.

4. Hider P, Griffin G, Walker M, Coughlan E. The information-seeking behavior of clinical staff in a large health care organization. Journal of the Medical Library Association : JMLA. 2009 Feb 1;97:47–50.

5. Cao Y, Liu F, Simpson P, Antieau L, Bennett A, Cimino JJ, et al. AskHERMES: An online question answering system for complex clinical questions. Journal of biomedical informatics. 2011;44(2):277–88.

6. Ben Abacha A, Zweigenbaum P. MEANS: A medical question-answering system combining NLP techniques and semantic Web technologies. Information Processing & Management. 2015 Sep;51(5):570–94.

7. Terol RM, Martínez-Barco P, Palomar M. A knowledge based method for the medical question answering problem. Comput Biol Med. 2007 Oct;37(10):1511–21.

8. Goodwin TR, Harabagiu SM. Medical Question Answering for Clinical Decision Support. In: Proceedings of the . ACM International Conference on Information & Knowledge Management ACM International Conference on Information and Knowledge Management. 2016. p. 297–306.

9. Ben Abacha A, Demner-Fushman D. A question-entailment approach to question answering. BMC Bioinformatics. 2019;20(1):511.

10. Zahid M, Mittal A, Joshi R, Atluri G. CLINIQA: A Machine Intelligence Based Clinical Question Answering System. 2018.

11. Demner-Fushman D, Lin J. Answering Clinical Questions with Knowledge-Based and Statistical Techniques. Computational Linguistics. 2007;33:63–103.

12. Cairns B, Nielsen RD, Masanz JJ, Martin JH, Palmer M, Ward W, et al. The MiPACQ clinical question answering system. AMIA . Annual Symposium proceedings AMIA Symposium. 2011;2011:171–80.

13. Niu Y, Hirst G, McArthur G, Rodriguez-Gianolli P. Answering Clinical Questions with Role Identification. In: Proceedings of the ACL 2003 Workshop on Natural Language Processing in Biomedicine - Volume 13 [Internet]. USA: Association for Computational

Linguistics; 2003. p. 73–80. (BioMed ’03). Available from: https://doi.org/10.3115/1118958.1118968

14. Brown TB, Mann B, Ryder N, Subbiah M, Kaplan J, Dhariwal P, et al. Language Models Are Few-Shot Learners. In: Proceedings of the 34th International Conference on Neural Information Processing Systems. Red Hook, NY, USA: Curran Associates Inc.; 2020. (NIPS’20).

15. Taylor R, Kardas M, Cucurull G, Scialom T, Hartshorn A, Saravia E, et al. Galactica: A Large Language Model for Science. 2022.

16. Sarrouti M, Ouatik El Alaoui S. SemBioNLQA: A semantic biomedical question answering system for retrieving exact and ideal answers to natural language questions. Artif Intell Med. 2020 Jan;102:101767.

17. Yu H, Lee M, Kaufman D, Ely J, Osheroff JA, Hripcsak G, et al. Development, implementation, and a cognitive evaluation of a definitional question answering system for physicians. Journal of biomedical informatics. 2007;40(3):236–51.

18. Kell G, Marshall I, Wallace B, Jaun A. What Would it Take to get Biomedical QA Systems into Practice? In: Proceedings of the 3rd Workshop on Machine Reading for Question Answering [Internet]. Punta Cana, Dominican Republic: Association for Computational Linguistics; 2021. p. 28–41. Available from: https://aclanthology.org/2021.mrqa-1.3

19. Athenikos SJ, Han H. Biomedical question answering: a survey. Comput Methods Programs Biomed. 2010;99(1):1–24.

20. Jin Q, Yuan Z, Xiong G, Yu Q, Ying H, Tan C, et al. Biomedical Question Answering: A Survey of Approaches and Challenges. ACM Comput Surv [Internet]. 2022 Jan;55(2). Available from: https://doi.org/10.1145/3490238

21. Popay J, Roberts H, Sowden A, Petticrew M, Arai L, Rodgers M, et al. Guidance on the conduct of narrative synthesis in systematic reviews: A product from the ESRC Methods Programme. 2006.

22. Wolff RF, Moons KGM, Riley RD, Whiting PF, Westwood M, Collins GS, et al. PROBAST: A Tool to Assess the Risk of Bias and Applicability of Prediction Model Studies. Ann Intern Med. 2019 Jan 1;170(1):51–8.

23. Kwong JCC, Khondker A, Lajkosz K, McDermott MBA, Frigola XB, McCradden MD, et al. APPRAISE-AI Tool for Quantitative Evaluation of AI Studies for Clinical Decision Support. JAMA Network Open. 2023 Sep;6(9):e2335377–e2335377.

24. Page MJ, McKenzie JE, Bossuyt PM, Boutron I, Hoffmann TC, Mulrow CD, et al. The PRISMA 2020 statement: an updated guideline for reporting systematic reviews. BMJ. 2021 Mar 29;372:n71.

25. Campbell M, McKenzie JE, Sowden A, Katikireddi SV, Brennan SE, Ellis S, et al. Synthesis without meta-analysis (SWiM) in systematic reviews: reporting guideline. BMJ [Internet]. 2020;368. Available from: https://www.bmj.com/content/368/bmj.l6890

26. Rakotoson L, Letaillieur C, Massip S, Laleye FAA. Extractive-Boolean Question Answering for Scientific Fact Checking. In: Proceedings of the 1st International Workshop on Multimedia AI against Disinformation [Internet]. New York, NY, USA: Association for Computing Machinery; 2022. p. 27–34. (MAD '22). Available from: https://doi.org/10.1145/3512732.3533580

27. Wu Y, Ting HF, Lam TW, Luo R. BioNumQA-BERT: Answering Biomedical Questions Using Numerical Facts with a Deep Language Representation Model. In: Proceedings of the 12th ACM Conference on Bioinformatics, Computational Biology, and Health Informatics [Internet]. New York, NY, USA: Association for Computing Machinery; 2021. (BCB '21). Available from: https://doi.org/10.1145/3459930.3469557

28. Tutos A, Mollá D. A Study on the Use of Search Engines for Answering Clinical Questions. In: Proceedings of the Fourth Australasian Workshop on Health Informatics and Knowledge Management - Volume 108. AUS: Australian Computer Society, Inc.; 2010. p. 61–8. (HIKM '10).

29. Ni Y, Zhu H, Cai P, Zhang L, Qui Z, Cao F. CliniQA : highly reliable clinical question answering system. Studies in health technology and informatics. 2012;180:215–9.

30. Vong W, Then PHH. Information seeking features of a PICO-based medical question-answering system. 2015 9th International Conference on IT in Asia (CITA). 2015;1–7.

31. Demner-Fushman D, Lin J. Situated Question Answering in the Clinical Domain: Selecting the Best Drug Treatment for Diseases. In: Proceedings of the Workshop on Task-Focused Summarization and Question Answering. USA: Association for Computational Linguistics; 2006. p. 24–31. (SumQA '06).

32. Alzubi JA, Jain R, Singh A, Parwekar P, Gupta M. COBERT: COVID-19 Question Answering System Using BERT. Arabian journal for science and engineering. 2021;1–11.

33. Francis N, Green A, Guagliardo P, Libkin L, Lindaaker T, Marsault V, et al. Cypher: An Evolving Query Language for Property Graphs. In: Proceedings of the 2018 International Conference on Management of Data [Internet]. New York, NY, USA: Association for Computing Machinery; 2018. p. 1433–45. (SIGMOD '18). Available from: https://doi.org/10.1145/3183713.3190657

34. Ozyurt IB, Bandrowski A, Grethe JS. Bio-AnswerFinder: a system to find answers to questions from biomedical texts. Database : the journal of biological databases and curation. 2020;2020.

35. Wu Y, Ting HF, Lam TW, Luo R. BioNumQA-BERT: Answering Biomedical Questions Using Numerical Facts with a Deep Language Representation Model. In Association for Computing Machinery; 2021. (BCB '21). Available from: https://doi.org/10.1145/3459930.3469557

36. Du Y, Pei B, Zhao X, Ji J. Deep scaled dot-product attention based domain adaptation model for biomedical question answering. Methods (San Diego, Calif). 2020;173:69–74.

37. Xu G, Rong W, Wang Y, Ouyang Y, Xiong Z. External features enriched model for biomedical question answering. BMC Bioinformatics. 2021 May 26;22(1):272.

38. I. B. Ozyurt, J. Grethe. Iterative Document Retrieval via Deep Learning Approaches for Biomedical Question Answering. In: 2019 15th International Conference on eScience (eScience). 2019. p. 533–8.

39. Zhang X, Jia Y, Zhang Z, Kang Q, Zhang Y, Jia H. Improving End-to-End Biomedical Question Answering System. In: Proceedings of the 8th International Conference on Computing and Artificial Intelligence [Internet]. New York, NY, USA: Association for Computing Machinery; 2022. p. 274–9. (ICCAI '22). Available from: https://doi.org/10.1145/3532213.3532254

40. Peng K, Yin C, Rong W, Lin C, Zhou D, Xiong Z. Named Entity Aware Transfer Learning for Biomedical Factoid Question Answering. IEEE/ACM Trans Comput Biol Bioinform. 2021 May 11;PP.

41. Zhu X, Chen Y, Gu Y, Xiao Z. SentiMedQAer: A Transfer Learning-Based Sentiment-Aware Model for Biomedical Question Answering. Front Neurorobot. 2022;16:773329.

42. Devlin J, Chang MW, Lee K, Toutanova K. BERT: Pre-training of Deep Bidirectional Transformers for Language Understanding. ArXiv. 2019;abs/1810.04805.

43. Lee J, Yoon W, Kim S, Kim D, Kim S, So CH, et al. BioBERT: a pre-trained biomedical language representation model for biomedical text mining. Bioinformatics. 2020 Feb 15;36(4):1234–40.

44. Radford A, Narasimhan K. Improving Language Understanding by Generative Pre-Training. In 2018. Available from: https://api.semanticscholar.org/CorpusID:49313245

45. Krallinger M, Krithara A, Nentidis A, Paliouras G, Villegas M. BioASQ at CLEF2020: Large-Scale Biomedical Semantic Indexing and Question Answering. In: Jose JM, Yilmaz E, Magalhães J, Castells P, Ferro N, Silva MJ, et al., editors. Advances in Information Retrieval. Cham: Springer International Publishing; 2020. p. 550–6.

46. Nentidis A, Krithara A, Bougiatiotis K, Paliouras G, Kakadiaris I. Results of the sixth edition of the BioASQ Challenge. In: Proceedings of the 6th BioASQ Workshop A challenge on large-scale biomedical semantic indexing and question answering [Internet]. Brussels, Belgium: Association for Computational Linguistics; 2018. p. 1–10. Available from: https://www.aclweb.org/anthology/W18-5301

47. Omar R, El-Makky N, Torki M. A Character Aware Gated Convolution Model for Cloze-style Medical Machine Comprehension. 2020 IEEE/ACS 17th International Conference on Computer Systems and Applications (AICCSA). 2020;1–7.

48. Yu H, Kaufman D. A cognitive evaluation of four online search engines for answering definitional questions posed by physicians. Pac Symp Biocomput. 2007;328–39.

49. Doucette JA, Khan A, Cohen R. A Comparative Evaluation of an Ontological Medical Decision Support System (OMeD) for Critical Environments. In: Proceedings of the 2nd ACM SIGHIT International Health Informatics Symposium [Internet]. New York, NY, USA: Association for Computing Machinery; 2012. p. 703–8. (IHI '12). Available from: https://doi.org/10.1145/2110363.2110444

50. Li Y, Yin X, Zhang B, Liu T, Zhang Z, Hao H. A Generic Framework for Biomedical Snippet Retrieval. 2015 3rd International Conference on Artificial Intelligence, Modelling and Simulation (AIMS). 2015;91–5.

51. Makar R, Kouta M, Badr A. A Service Oriented Architecture for Biomedical Question Answering System. 2008 IEEE Congress on Services Part II (services-2 2008). 2008;73–80.

52. Wen A, Elwazir MY, Moon S, Fan J. Adapting and evaluating a deep learning language model for clinical why-question answering. JAMIA open. 2020;3(1):16–20.

53. Tsatsaronis G, Balikas G, Malakasiotis P, Partalas I, Zschunke M, Alvers MR, et al. An overview of the BIOASQ large-scale biomedical semantic indexing and question answering competition. BMC bioinformatics. 2015;16:138.

54. Demner-Fushman D, Lin J. Answer Extraction, Semantic Clustering, and Extractive Summarization for Clinical Question Answering. In: Proceedings of the 21st International Conference on Computational Linguistics and the 44th Annual Meeting of the Association for Computational Linguistics [Internet]. USA: Association for Computational Linguistics; 2006. p. 841–8. (ACL-44). Available from: https://doi.org/10.3115/1220175.1220281

55. W. Weiming, D. Hu, M. Feng, L. Wenyin. Automatic Clinical Question Answering Based on UMLS Relations. In 2007. p. 495–8.

56. Pasche E, Teodoro D, Gobeill J, Ruch P, Lovis C. Automatic medical knowledge acquisition using question-answering. Studies in health technology and informatics. 2009;150:569–73.

57. Lee M, Cimino J, Zhu H, Sable C, Shanker V, Ely J, et al. Beyond Information Retrieval—Medical Question Answering. AMIA Annu Symp Proc. 2006 Feb;469–73.

58. Hristovski D, Dinevski D, Kastrin A, Rindflesch TC. Biomedical question answering using semantic relations. BMC bioinformatics. 2015;16(1):6.

59. Kaddari Z, Mellah Y, Berrich J, Bouchentouf T, Belkasmi MG. Biomedical Question Answering: A Survey of Methods and Datasets. 2020 Fourth International Conference On Intelligent Computing in Data Sciences (ICDS). 2020;1–8.

60. Singh Rawat BP, Li F, Yu H. Clinical Judgement Study using Question Answering from Electronic Health Records. Proceedings of machine learning research. 2019;106:216–29.

61. Cao YG, Ely J, Antieau L, Yu H. Evaluation of the Clinical Question Answering Presentation. In: Proceedings of the Workshop on Current Trends in Biomedical Natural Language Processing. USA: Association for Computational Linguistics; 2009. p. 171–8. (BioNLP ’09).

62. Jin ZX, Zhang BW, Fang F, Zhang LL, Yin XC. Health assistant: answering your questions anytime from biomedical literature. Bioinformatics (Oxford, England). 2019;35(20):4129–39.

63. Du Y, Pei B, Zhao X, Ji J. Hierarchical Multi-layer Transfer Learning Model for Biomedical Question Answering. 2018 IEEE International Conference on Bioinformatics and Biomedicine (BIBM). 2018;362–7.

64. Du Y, Guo W, Zhao Y. Hierarchical Question-Aware Context Learning with Augmented Data for Biomedical Question Answering. 2019 IEEE International Conference on Bioinformatics and Biomedicine (BIBM). 2019;370–5.

65. Mairittha T, Mairittha N, Inoue S. Improving Fine-Tuned Question Answering Models for Electronic Health Records. In: Adjunct Proceedings of the 2020 ACM International Joint Conference on Pervasive and Ubiquitous Computing and Proceedings of the 2020 ACM International Symposium on Wearable Computers [Internet]. New York, NY, USA: Association for Computing Machinery; 2020. p. 688–91. (UbiComp-ISWC ’20). Available from: https://doi.org/10.1145/3410530.3414436

66. M. Wasim, W. Mahmood, M. N. Asim, M. U. Khan. Multi-Label Question Classification for Factoid and List Type Questions in Biomedical Question Answering. IEEE Access. 2019;7:3882–96.

67. Ruan T, Huang Y, Liu X, Xia Y, Gao J. QAnalysis: a question-answer driven analytic tool on knowledge graphs for leveraging electronic medical records for clinical research. BMC Med Inform Decis Mak. 2019 Apr 1;19(1):82.

68. J. Qiu, Y. Zhou, Z. Ma, T. Ruan, J. Liu, J. Sun. Question Answering based Clinical Text Structuring Using Pre-trained Language Model. In: 2019 IEEE International Conference on Bioinformatics and Biomedicine (BIBM). 2019. p. 1596–600.

69. Gobeill J, Patsche E, Theodoro D, Veuthey A, Lovis C, Ruch P. Question answering for biology and medicine. 2009 9th International Conference on Information Technology and Applications in Biomedicine. 2009;1–5.

70. Sondhi P, Raj P, Kumar VV, Mittal A. Question processing and clustering in INDOC: a biomedical question answering system. EURASIP J Bioinform Syst Biol. 2007;2007(1):28576.

71. Olvera-Lobo MD, Gutiérrez-Artacho J. Question-answering systems as efficient sources of terminological information: an evaluation. Health Info Libr J. 2010 Dec;27(4):268–76.

72. B. Xu, H. Lin, B. Liu. Study on question answering system for biomedical domain. In: 2009 IEEE International Conference on Granular Computing. 2009. p. 626–9.

73. Cruchet S, Boyer C, van der Plas L. Trustworthiness and relevance in web-based clinical question answering. Studies in health technology and informatics. 2012;180:863–7.

74. Dimitriadis D, Tsoumakas G. Word embeddings and external resources for answer processing in biomedical factoid question answering. Journal of biomedical informatics. 2019;92:103118.

75. Bai J, Yin C, Zhang J, Wang Y, Dong Y, Rong W, et al. Adversarial Knowledge Distillation Based Biomedical Factoid Question Answering. IEEE/ACM Trans Comput Biol Bioinform. 2022 Mar 22;PP.

76. Naseem U, Dunn AG, Khushi M, Kim J. Benchmarking for biomedical natural language processing tasks with a domain specific ALBERT. BMC Bioinformatics. 2022 Apr 21;23(1):144.

77. Du Y, Yan J, Zhao Y, Lu Y, Jin X. Dual Model Weighting Strategy and Data Augmentation in Biomedical Question Answering. In: 2021 IEEE International Conference on Bioinformatics and Biomedicine (BIBM). 2021. p. 659–62.

78. Weinzierl MA, Harabagiu SM. Epidemic Question Answering: question generation and entailment for Answer Nugget discovery. J Am Med Inform Assoc. 2023 Jan 18;30(2):329–39.

79. Du Y, Yan J, Lu Y, Zhao Y, Jin X. Improving Biomedical Question Answering by Data Augmentation and Model Weighting. IEEE/ACM Trans Comput Biol Bioinform. 2022 Apr 29;PP.

80. Bai J, Yin C, Wu Z, Zhang J, Wang Y, Jia G, et al. Improving Biomedical ReQA With Consistent NLI-Transfer and Post-Whitening. IEEE/ACM Transactions on Computational Biology and Bioinformatics. 2022;1–12.

81. Yoon W, Jackson R, Lagerberg A, Kang J. Sequence tagging for biomedical extractive question answering. Bioinformatics. 2022 Aug 2;38(15):3794–801.

82. Pasche E, Teodoro D, Gobeill J, Ruch P, Lovis C. QA-driven guidelines generation for bacteriotherapy. AMIA Annu Symp Proc. 2009 Nov 14;2009:509–13.

83. Gobeill J, Gaudinat A, Pasche E, Vishnyakova D, Gaudet P, Bairoch A, et al. Deep Question Answering for protein annotation. Database : the journal of biological databases and curation. 2015;2015.

84. Raza S, Schwartz B, Ondrusek N. A Question-Answering System on COVID-19 Scientific Literature. In: 2022 IEEE 46th Annual Computers, Software, and Applications Conference (COMPSAC). 2022. p. 1331–6.

85. Kia MA, Garifullina A, Kern M, Chamberlain J, Jameel S. Adaptable Closed-Domain Question Answering Using Contextualized CNN-Attention Models and Question Expansion. IEEE Access. 2022;10:45080–92.

86. Raza S, Schwartz B, Rosella LC. CoQUAD: a COVID-19 question answering dataset system, facilitating research, benchmarking, and practice. BMC Bioinformatics. 2022 Jun 2;23(1):210.

87. Rajpurkar P, Zhang J, Lopyrev K, Liang P. SQuAD: 100,000+ Questions for Machine Comprehension of Text. 2016.

88. Rajpurkar P, Jia R, Liang P. Know What You Don’t Know: Unanswerable Questions for SQuAD. 2018.

89. Lin CY. ROUGE: A Package for Automatic Evaluation of Summaries. In: Text Summarization Branches Out [Internet]. Barcelona, Spain: Association for Computational Linguistics; 2004. p. 74–81. Available from: https://aclanthology.org/W04-1013

90. Papineni K, Roukos S, Ward T, Zhu WJ. Bleu: a Method for Automatic Evaluation of Machine Translation. In: Proceedings of the 40th Annual Meeting of the Association for Computational Linguistics [Internet]. Philadelphia, Pennsylvania, USA: Association for Computational Linguistics; 2002. p. 311–8. Available from: https://aclanthology.org/P02-1040

91. Oita M, Vani K, Oezdemir-Zaech F. Semantically Corroborating Neural Attention for Biomedical Question Answering. In: Cellier P, Driessens K, editors. Machine Learning and Knowledge Discovery in Databases. Cham: Springer International Publishing; 2020. p. 670–85.

92. Arabzadeh N, Bagheri E. A self-supervised language model selection strategy for biomedical question answering. Journal of Biomedical Informatics. 2023 Sep;146:104486.

93. Pergola G, Kochkina E, Gui L, Liakata M, He Y. Boosting Low-Resource Biomedical QA via Entity-Aware Masking Strategies. In: Proceedings of the 16th Conference of the European Chapter of the Association for Computational Linguistics: Main Volume [Internet]. Association for Computational Linguistics; 2021. Available from: https://doi.org/10.18653%2Fv1%2F2021.eacl-main.169

94. Yang Z, Zhou Y, Nyberg E. Learning to Answer Biomedical Questions: OAQA at BioASQ 4B. In: Proceedings of the Fourth BioASQ workshop [Internet]. Association for Computational Linguistics; 2016. Available from: https://doi.org/10.18653%2Fv1%2Fw16-3104

95. Krithara A, Nentidis A, Paliouras G, Kakadiaris I. Results of the 4th edition of BioASQ Challenge. In: Proceedings of the Fourth BioASQ workshop [Internet]. Association for Computational Linguistics; 2016. Available from: https://doi.org/10.18653%2Fv1%2Fw16-3101

96. Sarrouti M, Alaoui SOE. A Biomedical Question Answering System in BioASQ 2017. In: BioNLP 2017 [Internet]. Association for Computational Linguistics; 2017. Available from: https://doi.org/10.18653%2Fv1%2Fw17-2337

97. Nentidis A, Bougiatiotis K, Krithara A, Paliouras G, Kakadiaris I. Results of the fifth edition of the BioASQ Challenge - BioNLP 2017. BioNLP 2017 [Internet]. 2017; Available from: https://doi.org/10.18653%2Fv1%2Fw17-2306

98. Papagiannopoulou E, Papanikolaou Y, Dimitriadis D, Lagopoulos S, Tsoumakas G, Laliotis M, et al. Large-Scale Semantic Indexing and Question Answering in Biomedicine. In: Proceedings of the Fourth BioASQ workshop [Internet]. Association for Computational Linguistics; 2016. Available from: https://doi.org/10.18653%2Fv1%2Fw16-3107

99. Eckert F, Neves M. Semantic role labeling tools for biomedical question answering: a study of selected tools on the BioASQ datasets. In: Proceedings of the 6th BioASQ Workshop A challenge on large-scale biomedical semantic indexing and question answering [Internet]. Association for Computational Linguistics; 2018. Available from: https://doi.org/10.18653%2Fv1%2Fw18-5302

100. Shin HC, Zhang Y, Bakhturina E, Puri R, Patwary M, Shoeybi M, et al. BioMegatron: Larger Biomedical Domain Language Model. In: Proceedings of the 2020 Conference on Empirical Methods in Natural Language Processing (EMNLP) [Internet]. Association for Computational Linguistics; 2020. Available from: https://doi.org/10.18653%2Fv1%2F2020.emnlp-main.379

101. Nishida K, Nishida K, Yoshida S. Task-adaptive Pre-training of Language Models with Word Embedding Regularization. In: Findings of the Association for Computational Linguistics: ACL-IJCNLP 2021 [Internet]. Association for Computational Linguistics; 2021. Available from: https://doi.org/10.18653%2Fv1%2F2021.findings-acl.398

102. Jin Q, Dhingra B, Liu Z, Cohen W, Lu X. PubMedQA: A Dataset for Biomedical Research Question Answering. 2019. 2567 p.

103. Sarrouti M, Ouatik El Alaoui S. A passage retrieval method based on probabilistic information retrieval model and UMLS concepts in biomedical question answering. J Biomed Inform. 2017 Apr;68:96–103.

104. Brokos GI, Malakasiotis P, Androutsopoulos I. Using Centroids of Word Embeddings and Word Mover's Distance for Biomedical Document Retrieval in Question Answering. In: Proceedings of the 15th Workshop on Biomedical Natural Language Processing [Internet]. Association for Computational Linguistics; 2016. Available from: https://doi.org/10.18653%2Fv1%2Fw16-2915

105. Ozyurt IB. On the effectiveness of small, discriminatively pre-trained language representation models for biomedical text mining. In: Proceedings of the First Workshop on Scholarly Document Processing [Internet]. Association for Computational Linguistics; 2020. Available from: https://doi.org/10.18653%2Fv1%2F2020.sdp-1.12

106. Pappas D, Malakasiotis P, Androutsopoulos I. Data Augmentation for Biomedical Factoid Question Answering. In: Proceedings of the 21st Workshop on Biomedical Language Processing [Internet]. Association for Computational Linguistics; 2022. Available from: https://doi.org/10.18653%2Fv1%2F2022.bionlp-1.6

107. Wang XD, Leser U, Weber L. BEEDS: Large-Scale Biomedical Event Extraction using Distant Supervision and Question Answering. In: Proceedings of the 21st Workshop on Biomedical Language Processing [Internet]. Association for Computational Linguistics; 2022. Available from: https://doi.org/10.18653%2Fv1%2F2022.bionlp-1.28

108. Wang XD, Weber L, Leser U. Biomedical Event Extraction as Multi-turn Question Answering. In: Proceedings of the 11th International Workshop on Health Text Mining and Information Analysis [Internet]. Association for Computational Linguistics; 2020. Available from: https://doi.org/10.18653%2Fv1%2F2020.louhi-1.10

109. Neves M, Eckert F, Folkerts H, Uflacker M. Assessing the performance of Olelo, a real-time biomedical question answering application - BioNLP 2017. BioNLP 2017 [Internet]. 2017; Available from: https://doi.org/10.18653%2Fv1%2Fw17-2344

110. Wiese G, Weissenborn D, Neves M. Neural Domain Adaptation for Biomedical Question Answering - Proceedings of the 21st Conference on Computational Natural Language Learning (CoNLL 2017). Proceedings of the 21st Conference on Computational Natural Language Learning (CoNLL 2017) [Internet]. 2017; Available from: https://doi.org/10.18653%2Fv1%2Fk17-1029

111. Wiese G, Weissenborn D, Neves M. Neural Question Answering at BioASQ 5B - BioNLP 2017. BioNLP 2017 [Internet]. 2017; Available from: https://doi.org/10.18653%2Fv1%2Fw17-2309

112. Nishida K, Nishida K, Saito I, Asano H, Tomita J. Unsupervised Domain Adaptation of Language Models for Reading Comprehension - Proceedings of the Twelfth Language Resources and Evaluation Conference. Calzolari N, Béchet F, Blache P, Choukri K, Cieri C, Declerck T, et al., editors. Proceedings of the Twelfth Language Resources and Evaluation Conference. 2020 May;5392–9.

113. Yan Y, Zhang BW, Li XF, Liu Z. List-wise learning to rank biomedical question-answer pairs with deep ranking recursive autoencoders. PLoS One. 2020;15(11):e0242061.

114. Jin ZX, Zhang BW, Fang F, Zhang LL, Yin XC. A Multi-strategy Query Processing Approach for Biomedical Question Answering: USTB_PRIR at BioASQ 2017 Task 5B. In: BioNLP 2017 [Internet]. Association for Computational Linguistics; 2017. Available from: https://doi.org/10.18653%2Fv1%2Fw17-2348

115. Lee M, Cimino J, Zhu HR, Sable C, Shanker V, Ely J, et al. Beyond information retrieval–medical question answering. AMIA . Annual Symposium proceedings AMIA Symposium. 2006;2006:469–73.

116. Robles-Flores JA, Roussinov D. Examining Question-Answering Technology from the Task Technology Fit Perspective. Commun Assoc Inf Syst. 2012;30:26.

117. Law PM, Lo LYH, Endert A, Stasko J, Qu H. Causal Perception in Question-Answering Systems. In: Proceedings of the 2021 CHI Conference on Human Factors in Computing Systems [Internet]. New York, NY, USA: Association for Computing Machinery; 2021. Available from: https://doi.org/10.1145/3411764.3445444

118. Yang Z, Qi P, Zhang S, Bengio Y, Cohen W, Salakhutdinov R, et al. HotpotQA: A Dataset for Diverse, Explainable Multi-hop Question Answering. In: Proceedings of the 2018 Conference on Empirical Methods in Natural Language Processing [Internet]. Brussels, Belgium: Association for Computational Linguistics; 2018. p. 2369–80. Available from: https://aclanthology.org/D18-1259

119. Schuff H, Adel H, Vu NT. F1 is Not Enough! Models and Evaluation Towards User-Centered Explainable Question Answering. In: Proceedings of the 2020 Conference on Empirical Methods in Natural Language Processing (EMNLP) [Internet]. Online: Association for Computational Linguistics; 2020. p. 7076–95. Available from: https://aclanthology.org/2020.emnlp-main.575

120. Chowdhery A, Narang S, Devlin J, Bosma M, Mishra G, Roberts A, et al. PaLM: Scaling Language Modeling with Pathways. ArXiv. 2022;abs/2204.02311.

121. Singhal K, Azizi S, Tu T, Mahdavi SS, Wei J, Chung HW, et al. Large Language Models Encode Clinical Knowledge [Internet]. arXiv; 2022. Available from: https://arxiv.org/abs/2212.13138

122. Pal A, Umapathi LK, Sankarasubbu M. MedMCQA: A Large-scale Multi-Subject Multi-Choice Dataset for Medical domain Question Answering. In: Flores G, Chen GH, Pollard T, Ho JC, Naumann T, editors. Proceedings of the Conference on Health, Inference, and Learning [Internet]. PMLR; 2022. p. 248–60. (Proceedings of Machine Learning Research; vol. 174). Available from: https://proceedings.mlr.press/v174/pal22a.html

123. Shaib C, Li ML, Joseph S, Marshall IJ, Li JJ, Wallace BC. Summarizing, Simplifying, and Synthesizing Medical Evidence Using GPT-3 (with Varying Success). 2023.

124. Abimbola S, Patel B, Peiris D, Patel A, Harris M, Usherwood T, et al. The NASSS framework for ex post theorisation of technology-supported change in healthcare: worked example of the TORPEDO programme. BMC Medicine. 2019 Dec 30;17(1):233.

125. Greenhalgh T, Wherton J, Papoutsi C, Lynch J, Hughes G, A'Court C, et al. Beyond Adoption: A New Framework for Theorizing and Evaluating Nonadoption, Abandonment, and Challenges to the Scale-Up, Spread, and Sustainability of Health and Care Technologies. J Med Internet Res. 2017 Nov 1;19(11):e367.

126. Cimino J, Friedmann B, Jackson K, Li J, Pevzner J, Wrenn J. Redesign of the Columbia University Infobutton Manager. AMIA . Annual Symposium proceedings / AMIA Symposium AMIA Symposium. 2007 Feb 1;2007:135–9.

# Appendix

## SECTION A: DATA COLLECTION

The data collection form, piloted by GK, LQ and DF, was used to manually extract data from each included article. The variables extracted included publication year, author, journal title/conference name, article title, question answering domain, method and approach; source of questions for training and evaluation, source of answers for training and evaluation, form (i.e. factoid, definition, yes/no) of answers used for training and evaluation. We also evaluated the papers against the utility criteria outlined in [1], i.e. whether or not reliable sources of health information are used to derive the answers, whether or not the answers in the form of guidance, whether or not the answers useful In the context in which healthcare providers would be practicing, whether or not there is sufficient “rationale” for the answers provided, whether or not the system resolve conflicting evidence appropriately, whether or not the system handle and communicate uncertainties adequately, size of the training set, size of the evaluation set, quantitative results based on the training data, and quantitative results based on the evaluation data.

## SECTION B: CRITERIA RATINGS

*Table 2: Examples of texts that satisfy each criterion 'completely', 'partially' or 'not at all'.'*

| Domain | Examples that completely satisfy criterion | Example that partially satisfy criterion | Example that does not satisfy criterion at all |
|---|---|---|---|
| **Reliable sources** | Sources that were verified and approved by medical professionals, e.g. HON-certified health websites.<br><br>The QA system contains a component that rates the reliability of answer sources when a mixture of sources is used. | Sources that have not been described/mentioned in the corresponding studies<br><br>Biomedical databases (e.g. PubMed) that contain a mixture of reliable (e.g. randomized control trials) and unreliable (e.g. opinion article) sources | Databases and search systems that contain and retrieve information that is predominantly non-medical in nature and has not been verified by medical professionals, e.g. Google. |
| **Answers in the form of guidance** | Answers that suggest guidelines that may be relevant, e.g. "No relevant local or national guidelines are available, but here is one from Wirral Community Teaching Hospital".<br><br>Answers that provide conditional suggestions e.g. when recommending Nitrofurantoin: "If the estimated glomerular filtration rate (eGFR) ≥ 45 ml/minute then 100 mg modified-release twice a day (or if unavailable, 50 mg four times a day) for 3 days." | Answers consisting of extracted text spans, sentences or paragraphs from a particular text that may be in the form of guidance. | Answers are in the form of factoids (e.g. 100g of aspirin) or single words.<br><br>Answers provide definitive but unverifiable instructions (e.g. "Prescribe X medicine for Y condition"). |
| **Useful in the context in which the provider is practicing** | The answers are based on the location of the clinician, e.g. resources available to the hospital/clinic, antibiotic resistance. | Answers may account for one location-factor affecting the answer, e.g. antibiotic resistance, without considering any others. | The answers do not account for the situation in which the clinician is practicing. |
| **Sufficient "rationale" for the answers provided** | The system supports a particular answer either with additional text, source links/references or both. | Answers consisting of extracted text spans, sentences or paragraphs from a particular text that may contain rationale. | The answer does not contain any explanation or references. There is only a factoid, single word or phrase. |
| **Resolve conflicting evidence appropriately** | The system can identify conflicting evidence and communicate and communicate any conflicts to the clinicians, e.g. "3 systematic reviews were found, but their | The system identifies conflicting evidence and choose the most likely source without communicating the conflicts to the clinician. | The system assumes there is only one possible correct answer. |
| | conclusions are contradictory". | The system recognizes the conflicting evidence and synthesises it accordingly into an answer without informing the clinician of the original sources. | |
| **Handle and communicate uncertainties adequately** | The system communicates any sources of uncertainty and abstains from providing explicit guidance where appropriate. A good quality system would, for example, provide a caution if the answer came from low quality research evidence (a small or poor quality study). | The system identifies uncertain sources and excludes them from the answer but does not communicate their existence to the clinician. | The system is certain about every answer.<br><br>The system does not communicate uncertainties. |

# Section C: Types of systems

*Table 3: Grouping of papers according to system type.*

| System type | Papers |
|---|---|
| Knowledge graph | [2] |
| Neural | [3–29] |
| Modular | [2,4,5,11–13,17,19,22,23,27–60] |

*Table 4: Results of the BioASQ 5b challenge, including source citation, system type, name and average metrics from batches 1-5.*

| | | | Average results | | |
|---|---|---|---|---|---|
| **Paper** | **Type of system** | **Name of system** | **MRR** | **Lenient Accuracy** | **Strict Accuracy** |
| [9] | Modular, neural | HMQA | **0.50** | - | - |
| [48] | Modular, classical ML | OAQA | 0.22 | 0.30 | 0.16 |
| | Modular, classical ML | Proposed system | 0.25 | 0.29 | 0.22 |
| [12] | Modular, neural | fa1 | 0.33 | 0.45 | 0.26 |
| | Modular, neural | fa2 | 0.34 | 0.44 | 0.27 |
| | Modular, neural | fa1 (manually evaluated) | 0.41 | 0.55 | 0.34 |
| | Modular, neural | Deep QA | 0.46 | **0.56** | **0.39** |
| | Modular, rule-based | Lab Zhu, Fudan | 0.42 | 0.47 | 0.37 |

*Table 5: Results of the BioASQ 6b challenge, including source citation, system type, name and average metrics from batches 1-5.*

| | | | **Average results** | | |
|---|---|---|---|---|---|
| **Paper** | **Type of system** | **Name of system** | **MRR** | **Lenient Accuracy** | **Strict Accuracy** |
| [10] | Modular, neural | HMQA | 0.50 | 0.54 | 0.39 |
| | Modular, neural | HQACL | **0.57** | **0.69** | **0.42** |
| [12] | Modular, neural | fa1 | 0.25 | 0.36 | 0.18 |
| | Modular, neural | fa2 | 0.24 | 0.37 | 0.20 |
| | Modular, neural | fa3 | 0.24 | 0.37 | 0.21 |
| | Modular, rule-based | Lab Zhu, Fudan | 0.28 | 0.33 | 0.24 |
| | Modular, classical ML | OAQA | 0.21 | 0.26 | 0.17 |

# SECTION D: SOURCES OF TRAINING/EVALUATION QUESTION DATA

*Table 6: Grouping of papers according to question source.*

| **Question source** | **Papers** | **Number of papers** |
|---|---|---|
| Article titles and/or last sentences of abstracts | [3,20,25,61] | 4 |
| Physicians within clinical settings | [26,32,35,44,45,58] | 6 |
| Physicians who did not necessarily ask the questions in clinical settings | [30,31,36,39,41,50,51,53,55,59,62] | 11 |
| Expert panel | [3–5,7–13,15–24,26–29,34,37,46,48,52,60,62–78] | 46 |
| Health websites | [43,51] | 2 |
| Crowdworkers | [7,9,10,70,76,77] | 6 |
| Artificial conversations | [10] | 1 |
| Authors of the papers | [2,6,33,38,40,42,47,56,57,79,80] | 11 |
| Unclear sources | [14,39,54] | 3 |

With regards to sources of physicians' questions that may not have been asked in clinical settings, four of articles used the questions for user studies [30,41,45,55]. [31,36] had a high risk of bias and low applicability for the concern because although genuine physicians' questions were used, either specifically simple ones were selected or compound questions were simplified. [53] had both a high risk of bias and applicability concern because the questions were created by one of the authors who had medical qualifications and the questions were designed to match specific answers. The questions may therefore not have been as complex as those asked in clinical settings. Similarly, [50] had a high risk of bias and applicability as the questions were created by novice physicians specifically for the study and thus may not be reflective of what clinicians would ask in practice.

The following studies used questions created by the authors according to a template: [33,38,40,42,47,56,57]. All the studies had high applicability concerns and a high risk of bias, apart from [40,56] which had a low risk of bias. This is because the target application was a tool for medical knowledge acquisition for clinical decision support. Even though the authors of [6,51] did not employ a specific template while creating the questions, they were still unrealistically simple. Hence, the studies were deemed to be of high risk of bias and had high applicability concerns. Furthermore, [51] focused only on definitional questions. As [47] is purely a user study, it had no associated risk of bias or applicability assessment.

# SECTION E: EXAMPLE OF SUMMARY ANSWER FOR BioASQ

"When ciliary function is perturbed, photoreceptors may die, kidney tubules develop cysts, limb digits multiply and brains form improperly. Malformation of primary cilia in the collecting ducts of kidney tubules is accompanied by development of autosomal recessive polycystic kidney disease".

# SECTION F: SPECIALIZED QUESTION TOPICS

As the question and information sources of [50] were narrow, the risk of bias and applicability were high. All the drug-specific questions were also inapplicable to the review question for the same reason.

The questions used by [33,49,53] were deemed to be of high risk of bias because the research question of the paper exactly matched the review question. On the other hand, the systems developed in [40,56] addressed narrower research questions. Specifically, [40,56] aimed to create "machine-readable legacy knowledge rules" to generate guidelines for drug prescriptions.

# SECTION G: ANSWER SOURCES

*Table 7: Grouping of papers according to answer source.*

| Answer source | Papers | Number of papers |
|---|---|---|
| PubMed/MEDLINE | [30,2,31,61,32,34,37,38,40,41,4,42,5,7,44,45,8,3,46,9,10,47, 11,48–53,12,56,57,13–19,21–23,25,26,20,24,63–68,70,69,71–73,81,80,74–76,62,77,79,78,59,60,27–29] | 71 |
| Health websites | [43,51,55,58] | 4 |
| General QA websites | [36] | 1 |
| Online dictionaries | [32,41,51] | 3 |
| Preprints | [62] | 1 |
| Wikipedia | [7,9,10,30,35,70,76,77] | 8 |
| World Health Organisation | [62] | 1 |
| World wide web | [32,35,36,51,41,44] | 6 |
| Synthetic data | [33] | 1 |
| eMedicine documents | [30,45] | 2 |
| Clinical guidelines | [45] | 1 |
| Miscellaneous medical sources | [6,39,54] | 3 |

The limited control over the information contained in Wikipedia led to the answers derived using only Wikipedia [35] to be not at all reliable. Meanwhile, the answers that are derived from Wikipedia in tandem with other sources, e.g., biomedical databases and clinical notes, were partially reliable. The quality control over the information on Google is also limited, which is why answers derived only using Google are not at all reliable [36].

Additionally, general QA websites and miscellaneous medical sources were deemed to be partially reliable. This is because they were medical sources or, as in the case with the general QA websites, the answers to the questions were written by “topic experts”. The credentials of these “topic experts” are unknown. The synthetic data was not at all reliable as it may not be reflective of the real world.

# SECTION H: TYPES OF ANSWERS

*Table 8: Grouping of papers according to answer type.*

| Answer type | Papers | Number of papers |
|---|---|---|
| One word | [61] | 1 |
| Medical concepts | [2,63,67] | 3 |
| Definitions | [4,35–37,44] | 5 |
| Yes/no/unclear answers | [5,20,25,27,29,33,37,52,59,63,64,66,71,73,75,81] | 16 |
| Clustered answers | [30,38,39,45,47,50] | 6 |
| Factoids | [5,8–10,12,13,15–18,23,24,37,40,48,49,52,56,57,59,62–64,68–80] | 36 |
| Lists of factoids | [5,9,26,48,63,64,66,68–71,73] | 12 |
| Abstracts | [53] | 1 |
| Single sentence | [14,19,22,25,43,51,81] | 7 |
| Paragraphs/several sentences | [4,6,11,20,30,31,36–38,41,42,44–47,51,58,64,68,71,73,81] | 22 |
| Documents/webpages | [36,50,58,63,65,67,68] | 7 |
| URLs | [35,36] | 2 |
| Snippets | [28,34,46,60,63,75] | 6 |
| Unclear | [51,55] | 2 |

These studies had answers that were judged to be not relevant to clinical practice in the RoB assessment and QA criteria. The approaches described in the studies included one word answers [61] and lists of factoids. In [61], a cloze-style approach to question answering was applied. Under this setting, a word would be removed from a sentence and the system should then predict the missing word.

From the definitions, only the systems in [32,35] and [36] (Onelook) consist of absolute definitions and do not satisfy any criteria. [4,37,44] provided sentences which may contain guidance. Hence, they partially satisfy the guidance criterion. [54] (paragraphs, documents and webpages), [37] (ideal answers) and [4] (extracted sentences) may contain rationale, while the systems described in the other papers do not.

For the yes/no/unclear answer types, some may contain guidance due to accompanying sentences or paragraphs [37] or [51] (START). Due to the absolute nature of the answer type, the other systems do not contain guidance. [37] partially satisfies the criterion due to the existence of “ideal” answers which may contain rationale. [20] completely satisfies the rationale and conflict resolution criterion, as the yes/no/balanced/neutral answers are accompanied by context and conflicts are resolved by majority votes. The systems outlined in the other papers do not offer any rationale.

The sentences of extracts of documents [30,38,39,45,50] used for the clustered answers may contain rationales. On the other hand, abstracts [53] and single/multiple sentences [43] may contain guidance or rationales. The snippet-based answers all contain partial guidance, but only [46] provides rationales. All the documents and webpages, i.e. [36] (Google, PubMed, MedQA) and [50], may contain guidance and rationales.

Out of the paragraph answer type (total of 16), all systems apart from 2 may contain guidance ([30,45] only provide definitions). In addition, half of the systems may provide rationales (except [16,17,30,33,45,57,63,64]).

URL-based answers were not in the form of guidance and did not offer rationales, as they are only accompanied by strict definitions. Example responses were not included in [16,52][51] (MedQA) and [55] nor was the format of the answers described. Therefore, it is impossible to determine the reliability of the answers, as well as whether they contain answers.

# SECTION I: USABILITY

The only usability study was conducted by [47] which assessed the usability of the CliniCluster system. The system answers only therapy questions and presents the users with a hierarchy of interventions which are clustered by the I (intervention) and C (comparator) elements in a collection of documents. When a particular cluster is selected, the user is shown a ranked list answers tagged with P-O (probability-outcome) and I/C (intervention/comparator) elements. The usability was evaluated using a survey which was answered by 20 medical professionals. The participants examined the 25 questions included in CliniCluster before answering the survey. Aside from questions about the usability, the survey collected demographic information about the participants such as age, gender, years of clinical experience and medical specialty. Additionally, the survey asked participants to rate how familiar and difficult the therapy topics were to them.

# SECTION J: REFERENCES

1. Kell G, Marshall I, Wallace B, Jaun A. What Would it Take to get Biomedical QA Systems into Practice? In: Proceedings of the 3rd Workshop on Machine Reading for Question Answering [Internet]. Punta Cana, Dominican Republic: Association for Computational Linguistics; 2021. p. 28–41. Available from: https://aclanthology.org/2021.mrqa-1.3

2. Goodwin TR, Harabagiu SM. Medical Question Answering for Clinical Decision Support. In: Proceedings of the . ACM International Conference on Information & Knowledge Management ACM International Conference on Information and Knowledge Management. 2016. p. 297–306.

3. Wu Y, Ting HF, Lam TW, Luo R. BioNumQA-BERT: Answering Biomedical Questions Using Numerical Facts with a Deep Language Representation Model. In: Proceedings of the 12th ACM Conference on Bioinformatics, Computational Biology, and Health Informatics [Internet]. New York, NY, USA: Association for Computing Machinery; 2021. (BCB '21). Available from: https://doi.org/10.1145/3459930.3469557

4. Ozyurt IB, Bandrowski A, Grethe JS. Bio-AnswerFinder: a system to find answers to questions from biomedical texts. Database : the journal of biological databases and curation. 2020;2020.

5. Kaddari Z, Mellah Y, Berrich J, Bouchentouf T, Belkasmi MG. Biomedical Question Answering: A Survey of Methods and Datasets. 2020 Fourth International Conference On Intelligent Computing in Data Sciences (ICDS). 2020;1–8.

6. Alzubi JA, Jain R, Singh A, Parwekar P, Gupta M. COBERT: COVID-19 Question Answering System Using BERT. Arabian journal for science and engineering. 2021;1–11.

7. Du Y, Pei B, Zhao X, Ji J. Deep scaled dot-product attention based domain adaptation model for biomedical question answering. Methods (San Diego, Calif). 2020;173:69–74.

8. Xu G, Rong W, Wang Y, Ouyang Y, Xiong Z. External features enriched model for biomedical question answering. BMC Bioinformatics. 2021 May 26;22(1):272.

9. Du Y, Pei B, Zhao X, Ji J. Hierarchical Multi-layer Transfer Learning Model for Biomedical Question Answering. 2018 IEEE International Conference on Bioinformatics and Biomedicine (BIBM). 2018;362–7.

10. Du Y, Guo W, Zhao Y. Hierarchical Question-Aware Context Learning with Augmented Data for Biomedical Question Answering. 2019 IEEE International Conference on Bioinformatics and Biomedicine (BIBM). 2019;370–5.

11. I. B. Ozyurt, J. Grethe. Iterative Document Retrieval via Deep Learning Approaches for Biomedical Question Answering. In: 2019 15th International Conference on eScience (eScience). 2019. p. 533–8.

12. Dimitriadis D, Tsoumakas G. Word embeddings and external resources for answer processing in biomedical factoid question answering. Journal of biomedical informatics. 2019;92:103118.

13. Raza S, Schwartz B, Ondrusek N. A Question-Answering System on COVID-19 Scientific Literature. In: 2022 IEEE 46th Annual Computers, Software, and Applications Conference (COMPSAC). 2022. p. 1331–6.

14. Kia MA, Garifullina A, Kern M, Chamberlain J, Jameel S. Adaptable Closed-Domain Question Answering Using Contextualized CNN-Attention Models and Question Expansion. IEEE Access. 2022;10:45080–92.

15. Bai J, Yin C, Zhang J, Wang Y, Dong Y, Rong W, et al. Adversarial Knowledge Distillation Based Biomedical Factoid Question Answering. IEEE/ACM Trans Comput Biol Bioinform. 2022 Mar 22;PP.

16. Naseem U, Dunn AG, Khushi M, Kim J. Benchmarking for biomedical natural language processing tasks with a domain specific ALBERT. BMC Bioinformatics. 2022 Apr 21;23(1):144.

17. Raza S, Schwartz B, Rosella LC. CoQUAD: a COVID-19 question answering dataset system, facilitating research, benchmarking, and practice. BMC Bioinformatics. 2022 Jun 2;23(1):210.

18. Du Y, Yan J, Zhao Y, Lu Y, Jin X. Dual Model Weighting Strategy and Data Augmentation in Biomedical Question Answering. In: 2021 IEEE International Conference on Bioinformatics and Biomedicine (BIBM). 2021. p. 659–62.

19. Weinzierl MA, Harabagiu SM. Epidemic Question Answering: question generation and entailment for Answer Nugget discovery. J Am Med Inform Assoc. 2023 Jan 18;30(2):329–39.

20. Rakotoson L, Letaillieur C, Massip S, Laleye FAA. Extractive-Boolean Question Answering for Scientific Fact Checking. In: Proceedings of the 1st International Workshop on Multimedia AI against Disinformation [Internet]. New York, NY, USA: Association for Computing Machinery; 2022. p. 27–34. (MAD ’22). Available from: https://doi.org/10.1145/3512732.3533580

21. Du Y, Yan J, Lu Y, Zhao Y, Jin X. Improving Biomedical Question Answering by Data Augmentation and Model Weighting. IEEE/ACM Trans Comput Biol Bioinform. 2022 Apr 29;PP.

22. Bai J, Yin C, Wu Z, Zhang J, Wang Y, Jia G, et al. Improving Biomedical ReQA With Consistent NLI-Transfer and Post-Whitening. IEEE/ACM Transactions on Computational Biology and Bioinformatics. 2022;1–12.

23. Zhang X, Jia Y, Zhang Z, Kang Q, Zhang Y, Jia H. Improving End-to-End Biomedical Question Answering System. In: Proceedings of the 8th International Conference on Computing and Artificial Intelligence [Internet]. New York, NY, USA: Association for

Computing Machinery; 2022. p. 274–9. (ICCAI ’22). Available from: https://doi.org/10.1145/3532213.3532254

24. Peng K, Yin C, Rong W, Lin C, Zhou D, Xiong Z. Named Entity Aware Transfer Learning for Biomedical Factoid Question Answering. IEEE/ACM Trans Comput Biol Bioinform. 2021 May 11;PP.

25. Zhu X, Chen Y, Gu Y, Xiao Z. SentiMedQAer: A Transfer Learning-Based Sentiment-Aware Model for Biomedical Question Answering. Front Neurorobot. 2022;16:773329.

26. Yoon W, Jackson R, Lagerberg A, Kang J. Sequence tagging for biomedical extractive question answering. Bioinformatics. 2022 Aug 2;38(15):3794–801.

27. Oita M, Vani K, Oezdemir-Zaech F. Semantically Corroborating Neural Attention for Biomedical Question Answering. In: Cellier P, Driessens K, editors. Machine Learning and Knowledge Discovery in Databases. Cham: Springer International Publishing; 2020. p. 670–85.

28. Yan Y, Zhang BW, Li XF, Liu Z. List-wise learning to rank biomedical question-answer pairs with deep ranking recursive autoencoders. PLoS One. 2020;15(11):e0242061.

29. Arabzadeh N, Bagheri E. A self-supervised language model selection strategy for biomedical question answering. Journal of Biomedical Informatics. 2023 Sep;146:104486.

30. Cao Y, Liu F, Simpson P, Antieau L, Bennett A, Cimino JJ, et al. AskHERMES: An online question answering system for complex clinical questions. Journal of biomedical informatics. 2011;44(2):277–88.

31. Demner-Fushman D, Lin J. Answering Clinical Questions with Knowledge-Based and Statistical Techniques. Computational Linguistics. 2007;33:63–103.

32. Yu H, Kaufman D. A cognitive evaluation of four online search engines for answering definitional questions posed by physicians. Pac Symp Biocomput. 2007;328–39.

33. Doucette JA, Khan A, Cohen R. A Comparative Evaluation of an Ontological Medical Decision Support System (OMeD) for Critical Environments. In: Proceedings of the 2nd ACM SIGHIT International Health Informatics Symposium [Internet]. New York, NY, USA: Association for Computing Machinery; 2012. p. 703–8. (IHI ’12). Available from: https://doi.org/10.1145/2110363.2110444

34. Li Y, Yin X, Zhang B, Liu T, Zhang Z, Hao H. A Generic Framework for Biomedical Snippet Retrieval. 2015 3rd International Conference on Artificial Intelligence, Modelling and Simulation (AIMS). 2015;91–5.

35. Makar R, Kouta M, Badr A. A Service Oriented Architecture for Biomedical Question Answering System. 2008 IEEE Congress on Services Part II (services-2 2008). 2008;73–80.

36. Tutos A, Mollá D. A Study on the Use of Search Engines for Answering Clinical Questions. In: Proceedings of the Fourth Australasian Workshop on Health Informatics and Knowledge Management - Volume 108. AUS: Australian Computer Society, Inc.; 2010. p. 61–8. (HIKM ’10).

37. Tsatsaronis G, Balikas G, Malakasiotis P, Partalas I, Zschunke M, Alvers MR, et al. An overview of the BIOASQ large-scale biomedical semantic indexing and question answering competition. BMC bioinformatics. 2015;16:138.

38. Demner-Fushman D, Lin J. Answer Extraction, Semantic Clustering, and Extractive Summarization for Clinical Question Answering. In: Proceedings of the 21st International Conference on Computational Linguistics and the 44th Annual Meeting of the Association for Computational Linguistics [Internet]. USA: Association for Computational Linguistics; 2006. p. 841–8. (ACL-44). Available from: https://doi.org/10.3115/1220175.1220281

39. W. Weiming, D. Hu, M. Feng, L. Wenyin. Automatic Clinical Question Answering Based on UMLS Relations. In 2007. p. 495–8.

40. Pasche E, Teodoro D, Gobeill J, Ruch P, Lovis C. Automatic medical knowledge acquisition using question-answering. Studies in health technology and informatics. 2009;150:569–73.

41. Lee M, Cimino J, Zhu H, Sable C, Shanker V, Ely J, et al. Beyond Information Retrieval—Medical Question Answering. AMIA Annu Symp Proc. 2006 Feb;469–73.

42. Hristovski D, Dinevski D, Kastrin A, Rindflesch TC. Biomedical question answering using semantic relations. BMC bioinformatics. 2015;16(1):6.

43. Ni Y, Zhu H, Cai P, Zhang L, Qui Z, Cao F. CliniQA : highly reliable clinical question answering system. Studies in health technology and informatics. 2012;180:215–9.

44. Yu H, Lee M, Kaufman D, Ely J, Osheroff JA, Hripcsak G, et al. Development, implementation, and a cognitive evaluation of a definitional question answering system for physicians. Journal of biomedical informatics. 2007;40(3):236–51.

45. Cao YG, Ely J, Antieau L, Yu H. Evaluation of the Clinical Question Answering Presentation. In: Proceedings of the Workshop on Current Trends in Biomedical Natural Language Processing. USA: Association for Computational Linguistics; 2009. p. 171–8. (BioNLP ’09).

46. Jin ZX, Zhang BW, Fang F, Zhang LL, Yin XC. Health assistant: answering your questions anytime from biomedical literature. Bioinformatics (Oxford, England). 2019;35(20):4129–39.

47. Vong W, Then PHH. Information seeking features of a PICO-based medical question-answering system. 2015 9th International Conference on IT in Asia (CITA). 2015;1–7.

48. M. Wasim, W. Mahmood, M. N. Asim, M. U. Khan. Multi-Label Question Classification for Factoid and List Type Questions in Biomedical Question Answering. IEEE Access. 2019;7:3882–96.

49. Gobeill J, Patsche E, Theodoro D, Veuthey A, Lovis C, Ruch P. Question answering for biology and medicine. 2009 9th International Conference on Information Technology and Applications in Biomedicine. 2009;1–5.

50. Sondhi P, Raj P, Kumar VV, Mittal A. Question processing and clustering in INDOC: a biomedical question answering system. EURASIP J Bioinform Syst Biol. 2007;2007(1):28576.

51. Olvera-Lobo MD, Gutiérrez-Artacho J. Question-answering systems as efficient sources of terminological information: an evaluation. Health Info Libr J. 2010 Dec;27(4):268–76.

52. Sarrouti M, Ouatik El Alaoui S. SemBioNLQA: A semantic biomedical question answering system for retrieving exact and ideal answers to natural language questions. Artif Intell Med. 2020 Jan;102:101767.

53. Demner-Fushman D, Lin J. Situated Question Answering in the Clinical Domain: Selecting the Best Drug Treatment for Diseases. In: Proceedings of the Workshop on Task-Focused Summarization and Question Answering. USA: Association for Computational Linguistics; 2006. p. 24–31. (SumQA '06).

54. B. Xu, H. Lin, B. Liu. Study on question answering system for biomedical domain. In: 2009 IEEE International Conference on Granular Computing. 2009. p. 626–9.

55. Cruchet S, Boyer C, van der Plas L. Trustworthiness and relevance in web-based clinical question answering. Studies in health technology and informatics. 2012;180:863–7.

56. Pasche E, Teodoro D, Gobeill J, Ruch P, Lovis C. QA-driven guidelines generation for bacteriotherapy. AMIA Annu Symp Proc. 2009 Nov 14;2009:509–13.

57. Gobeill J, Gaudinat A, Pasche E, Vishnyakova D, Gaudet P, Bairoch A, et al. Deep Question Answering for protein annotation. Database : the journal of biological databases and curation. 2015;2015.

58. Cairns B, Nielsen RD, Masanz JJ, Martin JH, Palmer M, Ward W, et al. The MiPACQ clinical question answering system. AMIA . Annual Symposium proceedings AMIA Symposium. 2011;2011:171–80.

59. Ben Abacha A, Zweigenbaum P. MEANS: A medical question-answering system combining NLP techniques and semantic Web technologies. Information Processing & Management. 2015 Sep;51(5):570–94.

60. Sarrouti M, Ouatik El Alaoui S. A passage retrieval method based on probabilistic information retrieval model and UMLS concepts in biomedical question answering. J Biomed Inform. 2017 Apr;68:96–103.

61. Omar R, El-Makky N, Torki M. A Character Aware Gated Convolution Model for Cloze-style Medical Machine Comprehension. 2020 IEEE/ACS 17th International Conference on Computer Systems and Applications (AICCSA). 2020;1–7.

62. Pergola G, Kochkina E, Gui L, Liakata M, He Y. Boosting Low-Resource Biomedical QA via Entity-Aware Masking Strategies. In: Proceedings of the 16th Conference of the European Chapter of the Association for Computational Linguistics: Main Volume [Internet]. Association for Computational Linguistics; 2021. Available from: https://doi.org/10.18653%2Fv1%2F2021.eacl-main.169

63. Yang Z, Zhou Y, Nyberg E. Learning to Answer Biomedical Questions: OAQA at BioASQ 4B. In: Proceedings of the Fourth BioASQ workshop [Internet]. Association for Computational Linguistics; 2016. Available from: https://doi.org/10.18653%2Fv1%2Fw16-3104

64. Krithara A, Nentidis A, Paliouras G, Kakadiaris I. Results of the 4th edition of BioASQ Challenge. In: Proceedings of the Fourth BioASQ workshop [Internet]. Association for Computational Linguistics; 2016. Available from: https://doi.org/10.18653%2Fv1%2Fw16-3101

65. Brokos GI, Malakasiotis P, Androutsopoulos I. Using Centroids of Word Embeddings and Word Mover's Distance for Biomedical Document Retrieval in Question Answering. In: Proceedings of the 15th Workshop on Biomedical Natural Language Processing [Internet]. Association for Computational Linguistics; 2016. Available from: https://doi.org/10.18653%2Fv1%2Fw16-2915

66. Sarrouti M, Alaoui SOE. A Biomedical Question Answering System in BioASQ 2017. In: BioNLP 2017 [Internet]. Association for Computational Linguistics; 2017. Available from: https://doi.org/10.18653%2Fv1%2Fw17-2337

67. Jin ZX, Zhang BW, Fang F, Zhang LL, Yin XC. A Multi-strategy Query Processing Approach for Biomedical Question Answering: USTB_PRIR at BioASQ 2017 Task 5B. In: BioNLP 2017 [Internet]. Association for Computational Linguistics; 2017. Available from: https://doi.org/10.18653%2Fv1%2Fw17-2348

68. Neves M, Eckert F, Folkerts H, Uflacker M. Assessing the performance of Olelo, a real-time biomedical question answering application - BioNLP 2017. BioNLP 2017 [Internet]. 2017; Available from: https://doi.org/10.18653%2Fv1%2Fw17-2344

69. Wiese G, Weissenborn D, Neves M. Neural Domain Adaptation for Biomedical Question Answering - Proceedings of the 21st Conference on Computational Natural Language Learning (CoNLL 2017). Proceedings of the 21st Conference on Computational Natural Language Learning (CoNLL 2017) [Internet]. 2017; Available from: https://doi.org/10.18653%2Fv1%2Fk17-1029

70. Wiese G, Weissenborn D, Neves M. Neural Question Answering at BioASQ 5B - BioNLP 2017. BioNLP 2017 [Internet]. 2017; Available from: https://doi.org/10.18653%2Fv1%2Fw17-2309

71. Nentidis A, Bougiatiotis K, Krithara A, Paliouras G, Kakadiaris I. Results of the fifth edition of the BioASQ Challenge - BioNLP 2017. BioNLP 2017 [Internet]. 2017; Available from: https://doi.org/10.18653%2Fv1%2Fw17-2306

72. Papagiannopoulou E, Papanikolaou Y, Dimitriadis D, Lagopoulos S, Tsoumakas G, Laliotis M, et al. Large-Scale Semantic Indexing and Question Answering in Biomedicine. In: Proceedings of the Fourth BioASQ workshop [Internet]. Association for Computational Linguistics; 2016. Available from: https://doi.org/10.18653%2Fv1%2Fw16-3107

73. Eckert F, Neves M. Semantic role labeling tools for biomedical question answering: a study of selected tools on the BioASQ datasets. In: Proceedings of the 6th BioASQ Workshop A challenge on large-scale biomedical semantic indexing and question answering [Internet]. Association for Computational Linguistics; 2018. Available from: https://doi.org/10.18653%2Fv1%2Fw18-5302

74. Shin HC, Zhang Y, Bakhturina E, Puri R, Patwary M, Shoeybi M, et al. BioMegatron: Larger Biomedical Domain Language Model. In: Proceedings of the 2020 Conference on Empirical Methods in Natural Language Processing (EMNLP) [Internet]. Association for Computational Linguistics; 2020. Available from: https://doi.org/10.18653%2Fv1%2F2020.emnlp-main.379

75. Ozyurt IB. On the effectiveness of small, discriminatively pre-trained language representation models for biomedical text mining. In: Proceedings of the First Workshop on Scholarly Document Processing [Internet]. Association for Computational Linguistics; 2020. Available from: https://doi.org/10.18653%2Fv1%2F2020.sdp-1.12

76. Nishida K, Nishida K, Saito I, Asano H, Tomita J. Unsupervised Domain Adaptation of Language Models for Reading Comprehension - Proceedings of the Twelfth Language Resources and Evaluation Conference. Calzolari N, Béchet F, Blache P, Choukri K, Cieri C, Declerck T, et al., editors. Proceedings of the Twelfth Language Resources and Evaluation Conference. 2020 May;5392–9.

77. Nishida K, Nishida K, Yoshida S. Task-adaptive Pre-training of Language Models with Word Embedding Regularization. In: Findings of the Association for Computational Linguistics: ACL-IJCNLP 2021 [Internet]. Association for Computational Linguistics; 2021. Available from: https://doi.org/10.18653%2Fv1%2F2021.findings-acl.398

78. Pappas D, Malakasiotis P, Androutsopoulos I. Data Augmentation for Biomedical Factoid Question Answering. In: Proceedings of the 21st Workshop on Biomedical Language Processing [Internet]. Association for Computational Linguistics; 2022. Available from: https://doi.org/10.18653%2Fv1%2F2022.bionlp-1.6

79. Wang XD, Leser U, Weber L. BEEDS: Large-Scale Biomedical Event Extraction using Distant Supervision and Question Answering. In: Proceedings of the 21st Workshop on Biomedical Language Processing [Internet]. Association for Computational Linguistics; 2022. Available from: https://doi.org/10.18653%2Fv1%2F2022.bionlp-1.28

80. Wang XD, Weber L, Leser U. Biomedical Event Extraction as Multi-turn Question Answering. In: Proceedings of the 11th International Workshop on Health Text Mining and Information Analysis [Internet]. Association for Computational Linguistics; 2020. Available from: https://doi.org/10.18653%2Fv1%2F2020.louhi-1.10

81. Jin Q, Dhingra B, Liu Z, Cohen W, Lu X. PubMedQA: A Dataset for Biomedical Research Question Answering. 2019. 2567 p.